\documentclass[journal]{IEEEtran}

\usepackage[utf8]{inputenc}
\usepackage[T1]{fontenc}
\usepackage{textcomp}
\usepackage{graphicx}
\usepackage{booktabs}
\usepackage{array}
\usepackage{amsmath}
\usepackage{amssymb}
\usepackage{float}
\usepackage{cite}
\usepackage{xcolor}
\usepackage[numbers]{natbib}
\usepackage[hidelinks]{hyperref}

\usepackage[normalem]{ulem}
\definecolor{darkgreen}{rgb}{0,.4,0}
\definecolor{darkcyan}{rgb}{0,.4,.4}
\newcommand{\REMOVE}[1]%
{{\color{red}\sout{#1}}}
\newcommand{\ADD}[1]{{\color{blue}{#1}}}

\newcommand{\COMMENT}[1]%
{{\color{darkgreen}\textbf{{Swalpa: }} {#1}}}

\graphicspath{{figures/}}

\begin{document}

\title{Cryo-Bench: Benchmarking Geospatial Foundation Models for Cryosphere Mapping}

\author{Saurabh~Kaushik, 
        Lalit~Maurya, 
        Beth~Tellman, 
        Swalpa~Kumar~Roy,~\IEEEmembership{Senior Member,~IEEE,}
        Valerio~Marsocci, 
        Gustau Camps-Valls~\IEEEmembership{Fellow,~IEEE,}
        and Jocelyn Chanussot,~\IEEEmembership{Fellow,~IEEE}
\thanks{\textit{Corresponding author}: S. Kaushik}
\thanks{S. Kaushik and B. Tellman are with the University of Wisconsin--Madison, Madison, WI, USA (e-mail: skaushik8@wisc.edu; tellman@wisc.edu).}
\thanks{L. Maurya is with the School of Computing and Mathematics, Manchester Metropolitan University, Manchester, M1 5GD, UK (e-mail: l.maurya@mmu.ac.uk).}
\thanks{S. K. Roy is with the Department of Computer Science and Engineering, Tezpur University, Assam 784028, India (e-mail: swalpa@tezu.ernet.in).}
\thanks{V. Marsocci is with ESA, ESRIN, $\varphi$-lab, Frascati, Italy.(e-mail: valerio.marsocci@esa.int)}
\thanks{G. Camps-Valls is with Universitat de València, 46980 València, Spain (e-mail: gustau.camps@uv.es).}
\thanks{J. Chanussot is with INRIA, CNRS, Grenoble INP, LJK, the Univ. Grenoble Alpes, 38000 Grenoble, France (e-mail: jocelyn.chanussot@grenoble-inp.fr).}
}

\markboth{}{Kaushik \MakeLowercase{\textit{et al.}}: Cryo-Bench}

\maketitle

\begin{abstract}
Geo-Foundation Models (GFMs) have been evaluated across diverse Earth observation tasks and domains, showing strong potential to produce reliable maps even with sparse labels. However, systematic benchmarking of GFMs for Cryosphere applications remains limited, primarily because suitable evaluation datasets are scarce. We address this gap by introducing Cryo-Bench, a benchmark comprising six semantic segmentation datasets covering five cryospheric components: supraglacial debris, glacial lakes under two sensing configurations, sea ice, calving fronts and Antarctic ice-shelf extent. The benchmark includes multispectral, RGB, and synthetic aperture radar observations from regions underrepresented in existing pretraining archives. We evaluate thirteen GFMs alongside U-Net and Vision Transformer baselines trained from scratch under a unified evaluation protocol. With frozen encoders, the U-Net achieves the highest six-dataset average mean intersection over union (mIoU) of 69.31\%, exceeding TerraMind (67.86\%) by 1.45 points. The paired difference has a 95\% confidence interval of [+0.94, +1.98], indicating that the U-Net's lead is statistically significant. Fine-tuning with a fixed learning rate produces mixed outcomes, with leading models such as TerraMind and DOFA experiencing declines in average mIoU (-0.20 and -1.31).
In contrast, learning-rate optimization substantially improves fine-tuning performance: DOFA reaches 93.97\% mIoU on the RGB glacial lake task, ranking the U-Net fourth, while Scale-MAE and GFM-Swin surpass U-Net on calving fronts. Averaged across all six datasets, four GFMs, TerraMind, GFM-Swin, DOFA, and Scale-MAE, exceed the U-Net baseline (69.31\%). In the few-shot setting, five GFMs, DOFA, RemoteCLIP, TerraMind, GFM-Swin, and Scale-MAE, likewise outperform U-Net; averaged across all thirteen GFMs, retention is 92.5\% of full-label accuracy compared with 86.1\% for U-Net. Encoder parameter count shows no clear relationship with segmentation accuracy.
\end{abstract}


\IEEEpeerreviewmaketitle

\noindent\textbf{Highlights}
\begin{itemize}
\item Six cryosphere segmentation datasets over five components, released as one benchmark.
\item Thirteen foundation models compared with U-Net and ViT under a single protocol.
\item Bootstrap intervals show the from-scratch U-Net leads frozen encoders with statistical significance.
\item Few-shot experiments reveal an advantage of GFMs over training a CNN from scratch.
\item Per-model learning-rate selection moves foundation models ahead of the U-Net.
\end{itemize}

\section{Introduction}
The development of Geospatial Foundation Models (GFMs) trained on remote sensing data has paved the way for a new paradigm in Earth Observation \cite{xiong2024, szwarcman2025, janowicz2025, zhou2026}. GFMs have shifted Earth observation from task-specific networks to encoders pretrained once on large unlabelled datasets and reused across downstream problems \cite{janowicz2025,zhou2026}. GFMs can handle diverse sensor inputs, carry spatiotemporal encodings, and produce usable maps even when labels are scarce \cite{szwarcman2025,janowicz2025,zhou2026}. These models are rigorously evaluated on benchmarks such as PANGAEA \cite{marsocci2024} and GEO-Bench \cite{lacoste2023}, which includes datasets spanning forest monitoring, crop type mapping, and disaster response (e.g., floods, wildfires). In addition, these benchmarks provide reproducible evaluation protocols by fixing a decoder, an optimizer, and a set of downstream tasks to ensure encoder evaluation rather than whole-pipeline evaluation.

However, the evaluation of GFMs for the Cryosphere, the frozen component of the Earth system encompassing glaciers, sea ice, glacial lakes, ice caps, and permafrost, has not yet been explored. The Cryosphere presents distinct challenges compared to other land-cover classes, driven by its strong sensitivity to climate change and highly dynamic processes \cite{rounce2023,glambie2025}. Most widely used pretraining datasets, for example TerraMesh \cite{blumenstiel2025}, SSL4EO-S12 \cite{wang2023}, FLAIR \cite{garioud2023} and MMEarth \cite{nedungadi2024}, either exclude or barely sample Greenland, Antarctica, the Canadian and Russian Arctic, and the high mountains of Asia. A model may see bright surfaces without learning cryospheric mapping structures, depending on where snow and ice pixels appear, because it does not distinguish them from seasonal snow.

In addition, most popular evaluation frameworks (PANGAEA and Geo-Bench) do not include cryosphere-focused downstream tasks, except for the CaFFe \cite{gourmelon2022} dataset, which represents glacier calving fronts in recent Geo-Bench-2. This gap matters operationally. Field data in glacierised and polar regions is expensive and sparse, so label-efficient methods carry real value \cite{kaushik2022,kaushik2024}. Cryospheric processes also unfold over a range of timescales, such as multi-decadal thinning, glacier surges, and outburst floods \cite{dehecq2019,guo2023,taylor2023}, and mapping them stresses segmentation in ways different from conventional land-cover tasks. For example, Debris-covered ice is spectrally close to surrounding moraine. Sea-ice charts contain classes present in one or two percent of pixels. Calving fronts are narrow, and the radar imagery used to map them looks nothing like the optical data most encoders saw during pretraining. To evaluate existing GFM capabilities and their potential for domain adaptation to an Earth system component largely absent from pretraining data, we must evaluate them on Cryosphere applications.

We introduce Cryo-Bench and use it to pose five questions to bridge this gap. How well do frozen GFM features encode cryospheric structure? Does the ranking obtained with frozen encoders survive a change of regime? Do GFMs help when labels are scarce? Does the pretraining objective, pretraining modality, or encoder size predict performance? And what does a fair comparison cost once learning rates are selected per model and computation is measured consistently? To keep benchmark evaluation comparable, we provide splits and class statistics in full; per-class results accompany every aggregate. We select learning rates for each model based on validation rather than fixing them, evaluate whole scenes rather than just center crops, and associate uncertainty with the headline comparison. Section \ref{sec:4} sets out these choices, and Section \ref{sec:5} shows that three of them learning-rate selection, evaluation coverage, and input standardization, move outcomes by more than the variations between the majority of models.

Our contributions are outlined as follows:

\begin{enumerate}
\item Cryo-Bench, a six-dataset benchmark with complete split and class statistics that covers five cryospheric components across three sensing modalities.
\item A single evaluation protocol that includes full-scene evaluation, validation-based checkpoint and learning-rate selection, and reporting that takes class imbalance into account.
\item We compare thirteen GFMs to two from-scratch baselines under frozen, few-shot, and fine-tuned regimes; with bootstrap intervals attached to the headline table.
\item Evaluations of predictive performance including encoder size, pretraining objective, pretraining modality, and the sensitivity of outcomes to protocol choices.
\item Guidance for practitioners choosing a model under label and compute constraints, tied to measured cost.
\end{enumerate}

\section{Background}
\subsection{Geospatial foundation models}
Self-supervised learning (SSL) approaches have shifted GFMs from task and region-specific models to task and geography-agnostic representation learning~\cite{he2022,radford2021}. Progress in the geospatial domain has been strongly influenced by the success of SSL techniques, particularly masked autoencoders \cite{he2022} and contrastive learning developed on natural images \cite{radford2021,caron2021}. Consequently, the community introduced roughly 60 remote sensing vision GFMs between 2021 and 2025 \cite{zhou2026}, mostly based on Masked Autoencoder (MAE) or contrastive learning.

Since a comprehensive review of all GFMs is beyond the scope of this work, we provide a brief overview of several widely adopted, representative, and open-access models. The Prithvi series \cite{jakubik2023,szwarcman2025} was the first model to demonstrate MAE-based SSL for remote sensing tasks, using 3D positional encodings to capture spatial and temporal structure jointly. To enable multi-modality and input-agnostic inference, DOFA \cite{xiong2024} introduces dynamic patch embeddings that use each channel's central wavelength as a shared parameter across sensors. CROMA~\cite{fuller2023} combines optical and radar inputs through contrastive objectives. Other widely used SSL approaches based on contrastive learning are adopted by models such as Galileo \cite{tseng2025}, which prioritizes multi-modality and enriches global and local feature representations.

Another line of work focuses on self-distillation, as demonstrated by DINO \cite{caron2021}, which uses two different augmented views of the input processed by student and teacher encoders, where the student learns to replicate the teacher's outputs. DINO-MM extends this framework to multimodal remote sensing data and highlights the benefits of multimodal SSL methods. The success of vision-language models (VLMs) in the natural image domain has also motivated similar developments in the geospatial community. RemoteCLIP \cite{liu2024} is among the earliest remote-sensing-specific vision-language models and relies on aligned textual and visual embeddings.
TerraMind \cite{jakubik2025} can synthesize one modality from another by incorporating generative multimodal modeling. The four encoders in the SSL4EO-S12 family \cite{wang2023} vary only in their self-supervised objective, making them a useful internal control on objective selection. SatlasNet \cite{bastani2023} is trained with large-scale supervision rather than self-supervision, GFM-Swin \cite{mendieta2023} uses a hierarchical backbone, Scale-MAE \cite{reed2023} targets scale-aware representation, and SpectralGPT \cite{hong2024} models the spectral dimension explicitly.

\begin{figure*}[!t]
\centering
\includegraphics[width=0.8\textwidth]{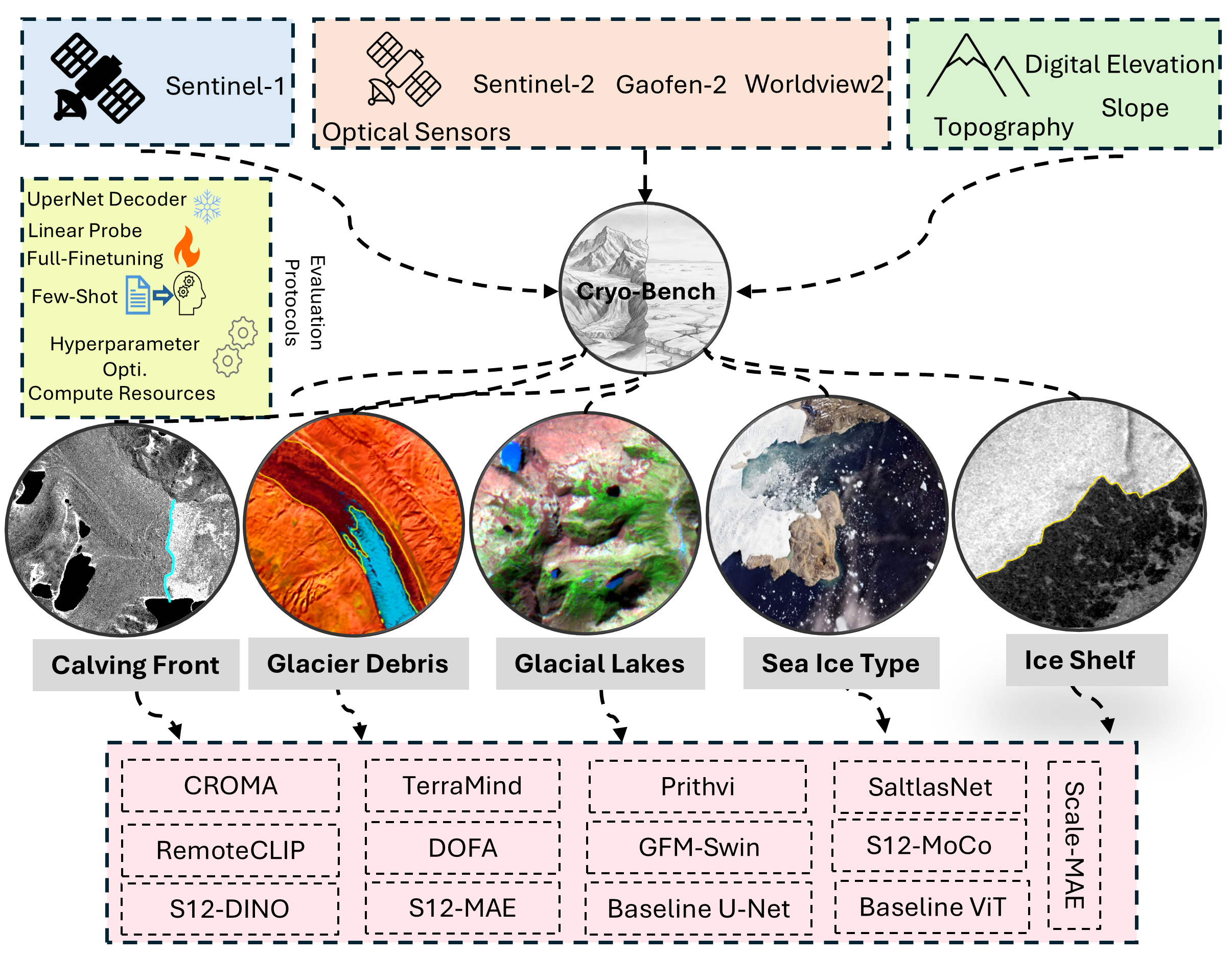}
\caption{Cryo-Bench covers six datasets and five cryospheric components across multispectral, RGB, and radar imagery. Every encoder is paired with the same UPerNet decoder and evaluated under the same three regimes. Dataset colors indicate the input modality.}
\label{fig:F1}
\end{figure*}

\subsection{Benchmarks for foundation models}
The rapid development of GFMs has motivated the creation of comprehensive evaluation datasets to assess model performance across diverse domains and sensing modalities. Among these, GEO-Bench~\cite{lacoste2023} and PANGAEA~\cite{marsocci2024} are the two widely used benchmarks covering urban mapping, agriculture, forestry, marine, and disaster management tasks (Table \ref{tab:T1}). SustainBench \cite{yeh2021} is another multimodal dataset oriented toward seven Sustainable Development Goals (SDGs) and incorporates data related to agriculture, health, education, and climate action. To ensure differences reflect representation rather than the training process, all three pair every encoder with a common decoder and a fixed optimization recipe.

Despite this progress, the Earth's Cryosphere, which spans environments from high-mountain glacier systems to the polar ice sheets, remains entirely overlooked in the assessment datasets currently in use. The Cryosphere is highly sensitive to climate change \cite{rounce2023,glambie2025} and responds across multiple timescales, for example, glacier ice flow changes \cite{dehecq2019}, glacial surge \cite{guo2023} and Glacial Lake Outburst Floods (GLOFs) \cite{taylor2023}. Due to mixed spectral responses, spectral similarity within classes, and substantial class imbalance in tasks involving small or narrow features, it is difficult to map key Cryosphere components from space, such as debris-covered glaciers, sea ice, calving fronts, and glacial lakes. Given these challenges, the large-scale pretraining and multiscale spatiotemporal representations learned by GFMs offer a promising opportunity to advance Cryosphere monitoring, especially in settings where labeled data is sparse.

\section{The Cryo-Bench benchmark}
\subsection{Selection criteria}
\begin{figure*}[!t]
\centering
\includegraphics[width=0.92\textwidth]{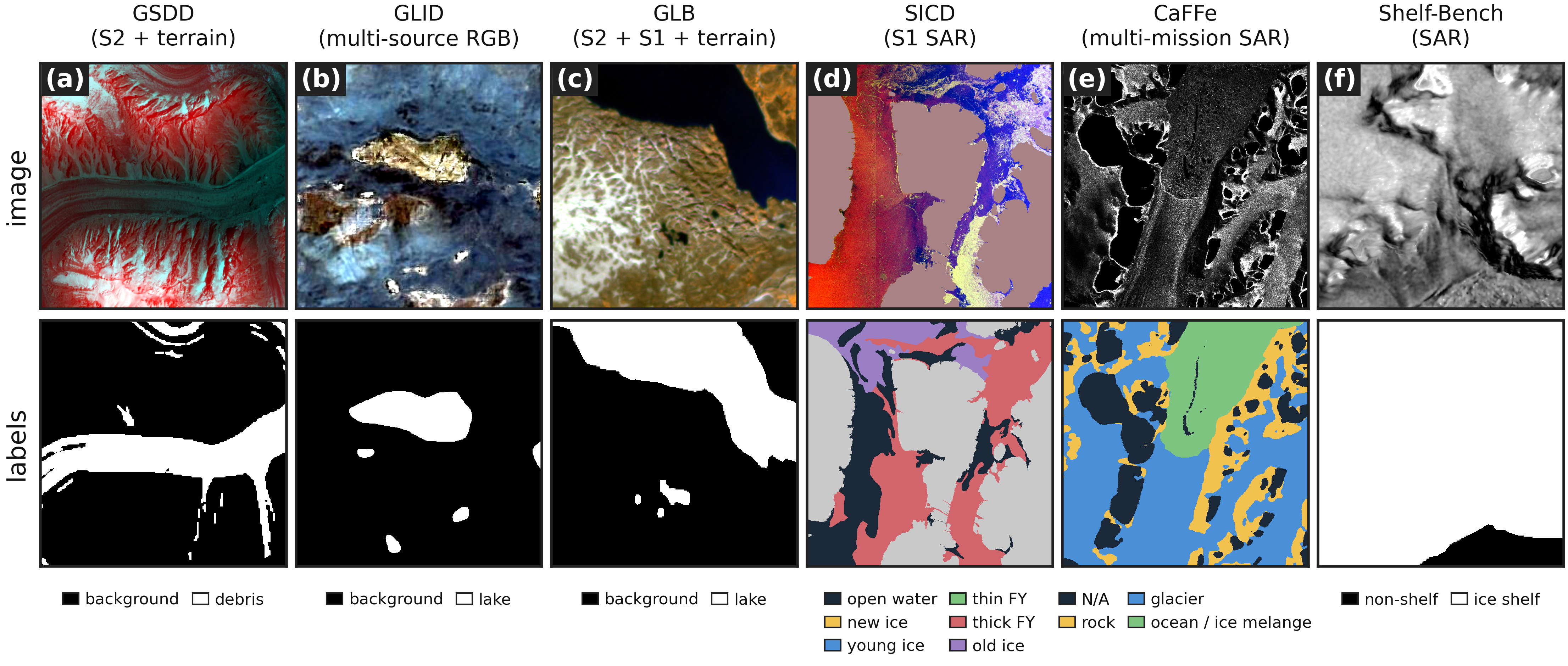}
\caption{Sample pairs of images and labels for each of the six Cryo-Bench datasets. When the dataset permitted, we chose scenes to show more than one class. When optical inputs make the target class more visible than true color, the inputs are shown as false-color composites: GSDD uses B5, B4, B3 and GLB uses B8, B4, B3 (it lacks B5); GLID is true color (B4, B3, B2), matching Figures 13 to 15. Binary tasks are drawn with a black background and a white target class, as in Figures \ref{fig:F13} to \ref{fig:F16}; multi-class tasks maintain the categorical palette and use grey marks to indicate pixels outside the labeled area/region.}
\label{fig:F2}
\end{figure*}

We assembled Cryo-Bench under five constraints: coverage of distinct cryospheric components, geographic spread, multiple sensing modalities, peer-reviewed provenance, and open access. The resulting six datasets cover five components (Table \ref{tab:T2}). An encoder trained only on optical imagery must transfer across modalities rather than merely across locations, since three are SAR-only. Figure \ref{fig:F2} shows sample labels and images.

\begin{table*}[!t]
\centering
\caption{Cryo-Bench compared with existing benchmarks used to evaluate geospatial foundation models.}
\label{tab:T1}
\footnotesize
\begin{tabular}{l>{\raggedright\arraybackslash}p{4.6cm}>{\raggedright\arraybackslash}p{4.2cm}>{\raggedright\arraybackslash}p{2.6cm}>{\raggedright\arraybackslash}p{2.2cm}}
\toprule
Benchmark & Domains / tasks & Task types & Modality & Temporality \\
\midrule
EarthNets & Land use / land cover & Classification, regression & MSI & Single \\
SustainBench & Sustainable-development goals & Classification, regression & MSI, street-view & Single \\
GEO-Bench & Urban, agriculture, forest & Classification, semantic segmentation & MSI & Single \\
PhilEO Bench & Urban & Classification, semantic segmentation & MSI & Single \\
FoMOBench & Urban, forest & Semantic segmentation & MSI & Single \\
PANGAEA & Urban, agriculture, forest, marine, disaster & Semantic segmentation, change detection, regression & MSI, SAR, DSM & Multi \\
Cryo-Bench (ours) & Cryosphere (debris, lakes, sea ice, calving fronts, ice shelves) & Semantic segmentation & MSI, RGB, SAR & Single \& multi \\
\bottomrule
\end{tabular}
\end{table*}

\begin{table*}[!t]
\centering
\caption{{Datasets associated with component, sensor, class count and split sizes in Cryo-Bench.}}
\label{tab:T2}
\footnotesize
\begin{tabular}{l>{\raggedright\arraybackslash}p{3.0cm}>{\raggedright\arraybackslash}p{3.6cm}rrrrr}
\toprule
Dataset & Component & Sensor & Classes & Train & Val & Test & 10\% tiles \\
\midrule
GSDD & Supraglacial debris & Sentinel-2 + terrain & 2 & 1,496 & 186 & 198 & \textasciitilde{}150 \\
GLID & Glacial lakes & Multi-source RGB & 2 & 14,400 & 1,600 & 2,367 & \textasciitilde{}1,440 \\
GLB & Glacial lakes & S2 + S1 + terrain (11 band) & 2 & 15,300 & 1,912 & 1,913 & \textasciitilde{}1,530 \\
SICD & Sea ice & Sentinel-1 SAR & 6 & 461 & 51 & 20 & \textasciitilde{}46 \\
CaFFe & Calving fronts & Multi-mission SAR & 4 & 504 & 55 & 122 & \textasciitilde{}50 \\
Shelf-Bench & Ice-shelf extent & SAR & 2 & 42,424 & 6,974 & 6,545 & \textasciitilde{}4,242 \\
\bottomrule
\end{tabular}
\end{table*}

\subsection{Datasets}
GSDD \cite{kaushik2025} uses global sampling to map supraglacial debris from Sentinel-2 with slope, elevation, and surface velocity as auxiliary bands. Debris and surrounding moraine are spectrally similar, making this a texture and context problem rather than a spectral one. Additionally, as seen in Section \ref{sec:5.1}, it is the benchmark's least discriminating task, with all models scoring within four points of one another.

GLID \cite{ma2025} uses three-band imagery from WorldView-2, Gaofen-2, Sentinel-2, and Landsat-8 to delineate glacial lakes across the Himalaya. Lakes occupy 1.9\% of test pixels, which is the most extreme foreground imbalance among the binary tasks in the benchmark.

The multi-source glacial lake dataset GLB \cite{kaushik2026} combines Sentinel-2, Sentinel-1, and terrain layers in eleven bands collected across the world's glacierized regions. GLID and GLB share the same goals, but their mapping and sensing configurations differ.

SICD \cite{stokholm2024} provides Sentinel-1 sea-ice charts over the Canadian and Greenlandic Arctic with six ice classes. With 461 training and 20 test scenes, it is the smallest dataset by scene count; nonetheless, the scenes are large, roughly 5200 by 5000 pixels. Three of its six classes occupy roughly one to two percent of the pixels each, while old ice (45.1\%) and open water (33.1\%) together make up the majority.

CaFFe \cite{gourmelon2022} contains calving-front zone maps from six SAR missions over Greenland, Alaska, and the Antarctic Peninsula. Its labels are ocean, ice melange, rock, glacier, and N/A (undefined region). Only the first three are surface classes; the N/A label denotes regions outside the mapped domain and is retained here because it is part of the distributed segmentation task and the released annotation. The most challenging scenario for optical-only encoders is their single-channel radar input.

Shelf-Bench \cite{baumhoer2026shelf} uses SAR to map the extent of the Antarctic ice shelf. As the only nearly balanced work in the benchmark at 48.7\% foreground, it adds the ice-shelf component and serves as a control against the more imbalanced lake tasks.

\subsection{Splits, class composition and what they imply}
We use the original partition of each source dataset. We use only the validation split for hyperparameter and checkpoint selection, while we evaluate the test split once for each configuration. Table \ref{tab:T2} shows split sizes, whereas Table \ref{tab:T3} shows class composition. Interpret the results based on three characteristics. First, from 1.9\% for GLID lakes to 48.7\% for ice shelves, the foreground proportion varies by more than an order of magnitude (Table \ref{tab:T3}). A model that predicts background nearly everywhere on the lake tasks still scores well on mean IoU, since the metric averages over classes. As a result, we present per-class IoU for the two multi-class tasks and foreground-class IoU for the four binary tasks.

Second, the number of tiles behind a 10\% few-shot experiment varies from roughly 46 for SICD to about 4,242 for Shelf-Bench, a difference of about two orders of magnitude. We compare label efficiency within a dataset and report retention rather than absolute few-shot scores when discussing the regime because few-shot difficulty is not similar across datasets. Third, scene size varies widely. GLB tiles are 256 pixels square, but SICD scenes are around 26 megapixels apiece. This is more than just a computational factor: it determines how much of a scene an evaluation process really scores, which proved to be a significant factor (Section \ref{sec:4}).

\begin{table}[!t]
\centering
\caption{Class composition of the test splits. Foreground share is 1.9\% on GLID and 4.5\% on GLB, against 48.7\% on Shelf-Bench.}
\label{tab:T3}
\footnotesize

\begin{tabular}{lp{5cm}}
\toprule
Dataset & Class composition of the test split \\
\midrule
GSDD & Debris 18.8\%, background 81.2\% \\
GLID & Lake 1.9\%, background 98.1\% \\
GLB & Lake 4.5\%, background 95.5\% \\
SICD & Old ice 45.1\%, open water 33.1\%, thick FY 16.8\%, young ice 2.3\%, thin FY 1.5\%, new ice 1.1\% \\
CaFFe & Glacier 49.4\%, rock 35.7\%, N/A 8.0\%, ocean/ice melange 6.9\% \\
Shelf-Bench & Ice shelf 48.7\%, non-shelf 51.3\% \\
\bottomrule
\end{tabular}%
\end{table}

\section{Models and evaluation protocol} \label{sec:4}
\subsection{Models}
In this study, we evaluate thirteen GFMs: CROMA, DOFA, GFM-Swin, Prithvi, RemoteCLIP, SatlasNet, Scale-MAE, SpectralGPT, the four SSL4EO-S12 encoders (MoCo, DINO, MAE, Data2Vec), and TerraMind. We train baseline models such as U-Net and a ViT from scratch using every accessible band. CROMA runs in its optical, joint, or SAR configuration with the dataset; thus, when a model comes in several variations, we select the one that matches the input. We classify encoders according to pretraining objective for the analyses in Sections \ref{subsec:5.6} and \ref{subsec:5.7} into masked image modeling (GFM-Swin, Scale-MAE, SpectralGPT, Prithvi, S12-MAE), contrastive (CROMA, S12-MoCo), self-distillation (S12-DINO, S12-Data2Vec), vision-language (RemoteCLIP), multimodal masked modeling (DOFA, TerraMind) and supervised (SatlasNet). Given that the four members of the SSL4EO-S12 family share an architecture and a pretraining corpus and differ only in aim, they are useful in this context.

\subsection{Training protocol}
We follow the PANGAEA protocol \cite{marsocci2024}. Each encoder is paired with a UPerNet decoder \cite{xiao2018} and optimized with AdamW (weight decay 0.05), batch size 8, cross-entropy loss, and multi-step decay at 60\% and 90\% of an 80-epoch schedule. Every five epochs, we run validation and transfer the checkpoint with the best validation mIoU to the test split.

Inputs are not resized to a common canvas. During training, the original tile is randomly cropped to match the encoder's native input size. Depending on the model, the crop size might vary between 120 and 224 pixels. By contrast, the U-Net processes the entire tile without cropping. 

\subsection{Regimes and learning-rate selection}
We evaluate GFMs primarily on three regimes. 1) Under frozen encoder, we fix the encoder and train only the decoder on the full training split. 2) Under few-shot, we again freeze the encoder, and the decoder sees a deterministic stratified 10\% sample. 3) Under fine-tuning, we train the encoder and decoder together while sweeping the learning rate over {10$^{-2}$, 10$^{-3}$, 10$^{-4}$, 10$^{-5}$} for fine-tuning, choose the rate that maximizes validation mIoU for each model, and report that configuration once on test. A set rate might yield better results for some encoders than others; reporting results at a fixed rate and results after tuning as two independent trials might confuse model quality with proximity to an arbitrary default. We trained all fifteen models at all four rates on all six datasets, so the validation-selected fine-tuning configuration is the fine-tuning result we report throughout (Table \ref{tab:T8}, Table \ref{tab:T8b}). We select based only on the validation data. For all models, the frozen and few-shot regimes continue to use a single decoder learning rate of 10$^{-4}$. 

\subsection{Band adaptation}
\label{subsec:IV-V}
Given different encoders accept different bands, a benchmark spanning optical and radar inputs must decide what each model receives  (Table \ref{tab:T4}). We pass the matching channels when dataset bands exactly match pretraining. Applying optical-only encoders to radar datasets results in a three-band proxy that replicates the radar channels. For radar-capable encoders, CaFFe supplies a single radar channel, and for optical encoders, this channel is repeated three times. On multi-band datasets, encoders restricted to three channels drop the terrain and derived layers they cannot ingest.

This creates an asymmetry favoring the baselines. U-Net and ViT use every available band on every dataset, while several GFMs observe only a proxy. Additionally, the U-Net also consumes the entire tile, whereas GFMs and the ViT can consider windows of 120 to 224 pixels. As a result, the U-Net has an advantage beyond computation on tasks where large-scale context is helpful. This is stated wherever baseline results are interpreted, and the value of radar pretraining on the radar tasks is quantified in Section \ref{subsec:5.7}.

\begin{table}[!t]
\centering
\caption{Band adaptation per dataset and encoder family.}
\label{tab:T4}
\footnotesize
\setlength{\tabcolsep}{3pt}
\begin{tabular}{>{\raggedright\arraybackslash}p{4.0cm}>{\raggedright\arraybackslash}p{4.0cm}}
\toprule
Situation & Handling \\
\midrule
Dataset bands match pretraining & Matching channels are used directly \\
Optical-only encoder on a SAR dataset & SAR replicated into a three-band proxy \\
Single-channel SAR (CaFFe), radar encoder & Fed directly \\
Single-channel SAR (CaFFe), optical encoder & Repeated three times \\
Multi-band dataset, three-channel encoder & Terrain and derived layers dropped \\
U-Net and ViT baselines & All available bands, full tile for U-Net \\
\bottomrule
\end{tabular}
\end{table}

\subsection{Metrics, uncertainty and reproducibility}
We report mIoU on the test set throughout. We use the validation set to select checkpoints and learning rates. We compute test scores with a sliding window over the entire scene rather than a single crop, averaging overlapping predictions. This protocol matters most for sea ice, whose roughly 5200 by 5000 pixel scenes are large enough that a single tile-sized crop would cover only 0.20\% to 1.03\% of the test pixels. In addition to mIoU, we also evaluated per-class IoU for the multi-class tasks and foreground-class IoU for the binary tasks. To quantify uncertainty, we evaluated each model on every dataset scene by scene and stored a per-scene confusion matrix. After bootstrapping, each of 2,000 replicates resamples scenes with replacement, sums their confusion matrices, and recomputes mIoU. Therefore, rather than applying to a related statistic, the interval applies to the specific data that are provided. Models that tile a dataset identically share replicates; therefore, such comparisons are aligned. This quantifies sensitivity to test-set composition. It does not capture training stochasticity, which requires multiple seeds and remains the best extension.

\section{Results} \label{sec:5}
\subsection{Frozen encoders} \label{sec:5.1}
\begin{figure*}[!t]
\centering
\includegraphics[width=0.92\textwidth]{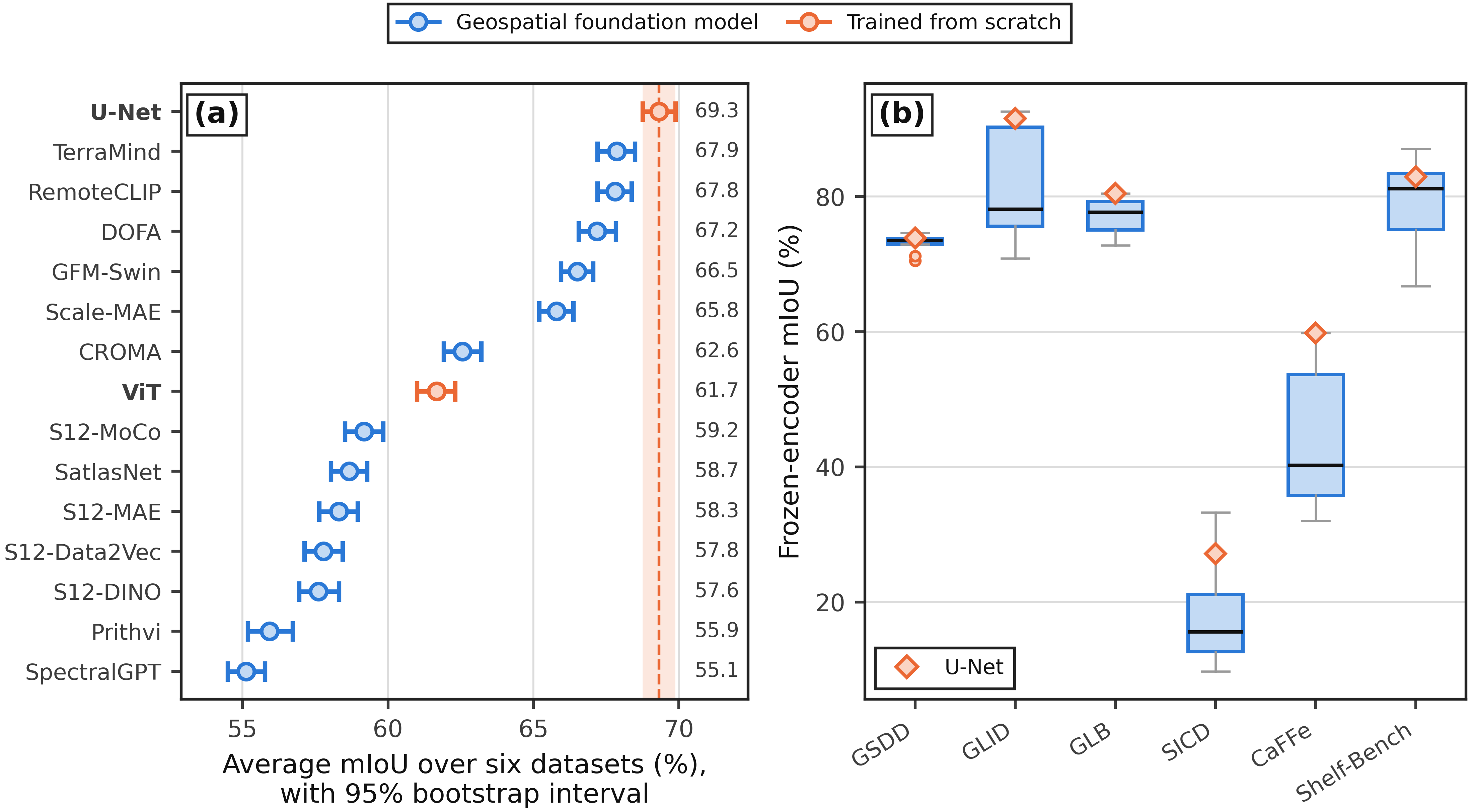}
\caption{Frozen-encoder results. (a) Six-dataset average mIoU, ranked, with 95\% bootstrap intervals from resampling test scenes; the shaded band and dashed line mark the leading interval and estimate, and TerraMind's interval does not overlap it, so the lead is resolved. (b) Distribution of those scores across the fifteen models within each dataset, showing how much each task separates representations overall.}
\label{fig:F3}
\end{figure*}

In the frozen encoder setting, the U-Net trained from scratch records the highest average of 69.31\% mIoU over the six datasets, followed by TerraMind at 67.86\% and RemoteCLIP at 67.81\% (Table \ref{tab:T5}, Figure \ref{fig:F3}a). The gap between the first and second scores is 1.45 points. A paired difference of +1.46 with a 95\% interval of [+0.94, +1.98] is obtained from bootstrapping test scenes; the U-Net leads in 100.0\% of replicates, and the interval excludes zero. As a result, the U-Net's lead over TerraMind is resolved under resampling of the test scenes, though the gap remains small relative to the protocol effects discussed in Section \ref{sec:5.10}. This is shown in Figure \ref{fig:F3}a, where the intervals of most neighboring pairs lower in the ranking still overlap and the interval widths for the six-dataset average range from 1.11 to 1.54 points across models.

Additionally, averages also conceal inconsistent performance across the tasks. U-Net outperforms on two of six datasets: multi-source glacial lakes (80.46\%) and calving fronts (59.82\%). TerraMind leads on debris cover (74.63\%) and sea ice (33.27\%); DOFA leads on the RGB lake task (92.61\%) and ice-shelf extent (87.08\%). The per-dataset columns in Table \ref{tab:T5} provide more information than the average for a practitioner selecting a model for a single application.

The degree to which the datasets differentiate models varies significantly (Figure \ref{fig:F3}b). Debris mapping hardly distinguishes between representations because the fifteen models on GSDD span just 4.1 points, from 70.52\% to 74.63\%. Calving fronts and sea ice span by 27.8 and 27.3 points, respectively. We therefore report per-dataset findings alongside any headline statistic, since the discriminating ones will dominate a benchmark average across both task types while the rest contribute equally to the mean.

Contrary to pretraining provenance, cross-modality transfer is more robust. Pretrained on RGB images without radar and polar coverage, Scale-MAE and GFM-Swin achieve 58.19\% and 58.12\% on calving fronts, respectively. They outperform radar-capable DOFA (at 50.71\%) and TerraMind (at 46.64\%). Radar backscatter is more accurately represented by whatever these encoders have learned about texture and edge structure than would be expected based on sensor provenance. This is extensively examined in Section \ref{subsec:5.7}. For all models, sea ice is the most challenging task. With an exceptionally large margin of 6.10 points, TerraMind achieves 33.27\%, while U-Net comes in second at 27.17\%. The contents of the numbers are examined in Section \ref{subsec:V-H}.

\begin{table*}[!t]
\centering
\caption{Frozen-encoder results with 100\% of the available labels, reported as test mIoU (\%). Baselines are trained from scratch. Avg Rank is the mean of a model's rank (1--15) on each of the six datasets individually, and can differ from the ordering by Avg mIoU.}
\label{tab:T5}
\footnotesize
\begin{tabular}{lrrrrrrrr}
\toprule
Model & GSDD & GLID & GLB & SICD & CaFFe & Shelf-Bench & Avg mIoU & Avg Rank \\
\midrule
U-Net & 73.89 & 91.58 & 80.46 & 27.17 & 59.82 & 82.97 & 69.31 & 2.67 \\
TerraMind & 74.63 & 88.26 & 79.91 & 33.27 & 46.64 & 84.46 & 67.86 & 3.17 \\
RemoteCLIP & 73.42 & 90.89 & 78.09 & 24.25 & 56.64 & 83.59 & 67.81 & 5.17 \\
DOFA & 72.96 & 92.61 & 79.39 & 20.41 & 50.71 & 87.08 & 67.19 & 4.83 \\
GFM-Swin & 73.00 & 89.69 & 78.42 & 14.17 & 58.12 & 85.63 & 66.50 & 6.33 \\
Scale-MAE & 73.47 & 90.91 & 79.16 & 9.74 & 58.19 & 83.28 & 65.79 & 6.33 \\
CROMA & 74.15 & 78.52 & 77.68 & 21.81 & 42.03 & 81.15 & 62.56 & 6.17 \\
ViT & 74.41 & 78.14 & 79.81 & 15.58 & 40.25 & 81.86 & 61.67 & 6.00 \\
S12-MoCo & 73.03 & 75.51 & 75.45 & 19.86 & 36.21 & 74.97 & 59.17 & 10.00 \\
SatlasNet & 73.70 & 77.03 & 74.74 & 19.62 & 33.96 & 73.00 & 58.67 & 10.17 \\
S12-MAE & 73.51 & 75.71 & 75.43 & 12.28 & 36.99 & 75.93 & 58.31 & 9.67 \\
S12-Data2Vec & 73.68 & 75.19 & 74.59 & 11.49 & 35.96 & 75.79 & 57.78 & 11.33 \\
S12-DINO & 71.19 & 75.69 & 75.17 & 12.74 & 35.58 & 75.28 & 57.61 & 11.67 \\
Prithvi & 70.52 & 71.11 & 74.96 & 12.58 & 32.01 & 74.41 & 55.93 & 13.50 \\
SpectralGPT & 73.22 & 70.87 & 72.82 & 14.39 & 32.70 & 66.75 & 55.12 & 13.00 \\
\bottomrule
\end{tabular}
\end{table*}

\subsection{Label efficiency}
\label{subsec:fewshot}
\begin{figure*}[!t]
\centering
\includegraphics[width=0.92\textwidth]{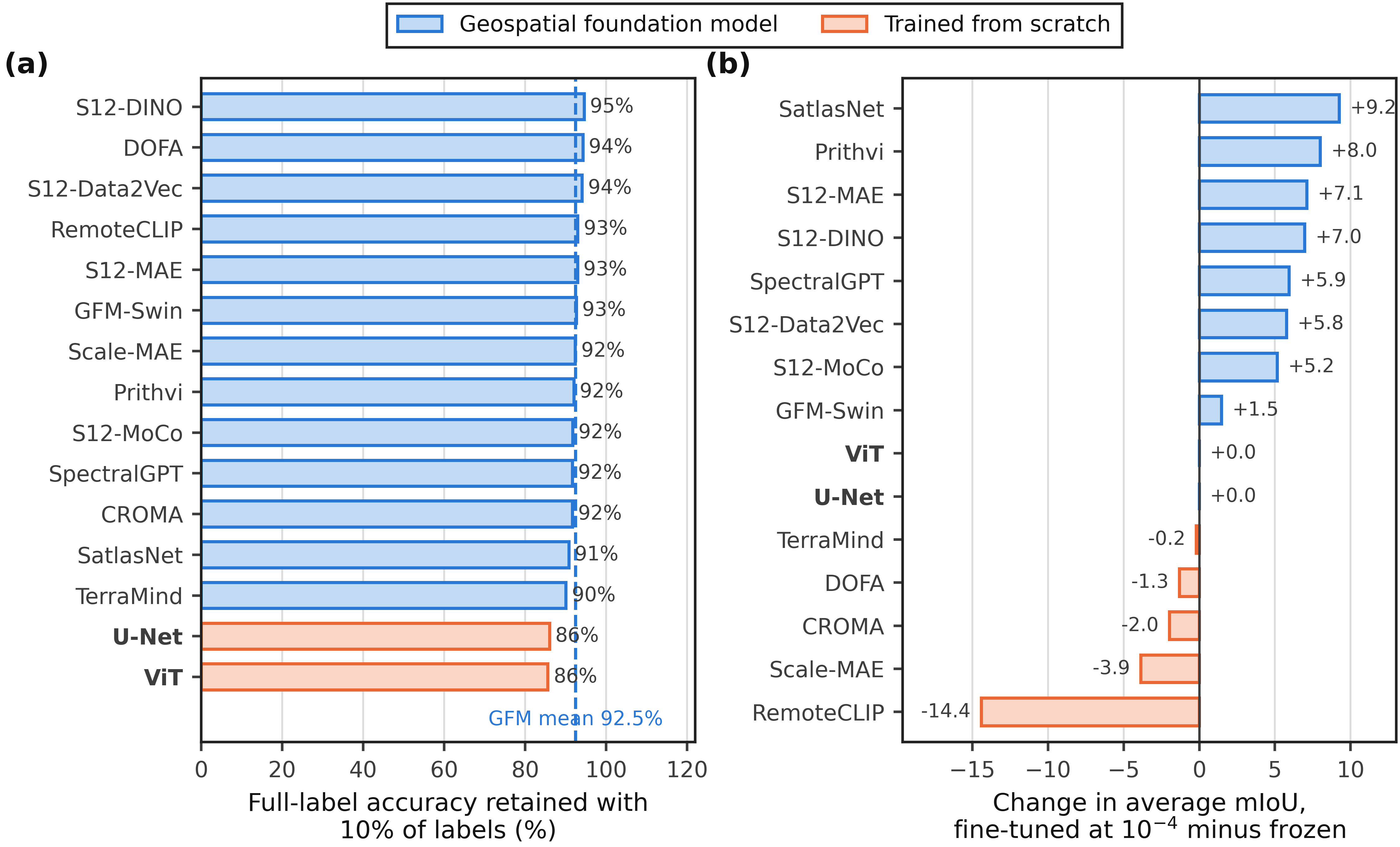}
\caption{Model robustness under changing evaluation regimes, measured relative to each model’s frozen performance: (a) the proportion of full-label accuracy retained when only 10\% of the labels are available; and (b) the change in average mIoU after fine-tuning with the fixed default learning rate of $10^{-4}$. Since the baseline models do not have a frozen–fine-tuned distinction, they are assigned a value of zero in (b).}
\label{fig:F4}
\end{figure*}

The few-shot experiment with 10\% of the labels reveals changes in ordering (Table \ref{tab:T6}). DOFA leads at 63.35\%, followed by RemoteCLIP at 63.08\% and GFM-Swin at 61.63\%. A further observation is that five GFMs, DOFA, RemoteCLIP, TerraMind, GFM-Swin, and Scale-MAE, rank higher than the U-Net, which drops to sixth place at 59.65\% (ranked by Avg mIoU throughout this section; Table \ref{tab:T6}'s Avg Rank column, the mean of each model's rank across the six datasets individually, need not agree). The advantage becomes clearer in terms of retention: when nine-tenths of the labels are removed, GFMs maintain 92.5\% of their full-label accuracy when averaged over the benchmark, compared to 86.1\% for the U-Net and 85.6\% for the ViT (Figure \ref{fig:F4}a), whereas S12-DINO retains 94.6\%. These observations are relevant for operational cryospheric applications, such as regular monitoring of sea ice, dangerous glacial lakes, or calving fronts, since obtaining high-quality labeled data is a persistent challenge and requires substantial human labor. Thus, the ability of GFMs to surpass the conventional U-Net carries direct operational significance. The ranking correlates at Spearman $\rho$ = 0.81 (p $<$ 0.001) between frozen and few-shot orderings.

\begin{table*}[!t]
\centering
\caption{Few-shot results with 10\% of labels and frozen encoders. Avg Rank is defined as in Table \ref{tab:T5}.}
\label{tab:T6}
\footnotesize
\begin{tabular}{lrrrrrrrr}
\toprule
Model & GSDD & GLID & GLB & SICD & CaFFe & Shelf-Bench & Avg mIoU & Avg Rank \\
\midrule
DOFA & 71.36 & 88.22 & 73.12 & 13.67 & 47.22 & 86.50 & 63.35 & 4.33 \\
RemoteCLIP & 65.56 & 86.74 & 70.92 & 19.48 & 55.00 & 80.80 & 63.08 & 5.83 \\
GFM-Swin & 64.86 & 84.79 & 71.19 & 13.89 & 53.04 & 82.01 & 61.63 & 7.17 \\
TerraMind & 69.90 & 79.82 & 74.88 & 21.44 & 38.15 & 82.67 & 61.14 & 5.00 \\
Scale-MAE & 66.40 & 84.86 & 71.25 & 8.93 & 53.06 & 80.29 & 60.80 & 7.17 \\
U-Net & 72.00 & 81.42 & 75.42 & 14.18 & 36.48 & 78.40 & 59.65 & 4.83 \\
CROMA & 71.75 & 71.48 & 72.83 & 16.79 & 33.52 & 77.78 & 57.36 & 6.33 \\
S12-DINO & 71.96 & 66.41 & 68.72 & 23.72 & 25.26 & 71.02 & 54.51 & 8.17 \\
S12-Data2Vec & 70.16 & 66.91 & 68.68 & 22.75 & 26.36 & 71.41 & 54.38 & 9.00 \\
S12-MoCo & 69.23 & 66.99 & 69.78 & 24.16 & 25.32 & 70.22 & 54.29 & 9.17 \\
S12-MAE & 69.32 & 67.28 & 68.57 & 22.94 & 25.13 & 72.12 & 54.23 & 9.17 \\
SatlasNet & 70.60 & 68.93 & 68.06 & 17.36 & 26.89 & 67.98 & 53.30 & 9.50 \\
ViT & 70.89 & 67.24 & 73.88 & 11.41 & 22.87 & 70.55 & 52.81 & 9.67 \\
Prithvi & 64.97 & 63.51 & 66.89 & 21.60 & 24.12 & 67.91 & 51.50 & 12.83 \\
SpectralGPT & 70.54 & 65.74 & 67.27 & 13.66 & 29.64 & 56.42 & 50.55 & 11.83 \\
\bottomrule
\end{tabular}
\end{table*}

\subsection{Fine-tuning at a fixed learning rate}
We observe that fine-tuning GFMs at the default 10$^{-4}$ leaves the U-Net first at 69.31\%, ahead of GFM-Swin at 67.97\% and SatlasNet at 67.92\% (Table \ref{tab:T7}). A quick reading of this fixed-rate fine-tuning experiment would support the conclusion that pretrained encoders do not overtake a well-trained convolutional neural network baseline on cryospheric tasks. Section \ref{subsec:V-D} shows why that conclusion is premature: with the learning rate selected per model and task rather than fixed at this default, the ordering changes substantially on every dataset we have swept.

The regime also reshuffles models relative to the frozen setting (Figure 4b). RemoteCLIP loses 14.41 points, the largest movement in the benchmark, while SatlasNet gains 9.24 and Prithvi 8.00. A compensation pattern runs through the regime: the five strongest frozen models lose an average of 2.89 points when fine-tuned at the fixed rate, while the five weakest gain 6.76, a ratio of 2.34. Weaker frozen representations have more headroom under end-to-end training, and stronger ones can be degraded by it. These observations highlight the importance of a lower learning rate when fine-tuning foundation models (Table \ref{tab:T8}).

\begin{table*}[!t]
\centering
\caption{Fine-tuning at a fixed learning rate of $10^{-4}$, on all six datasets. This is the sensitivity reference against which per-model learning-rate selection (Table \ref{tab:T8}, Table \ref{tab:T8b}) is compared. Avg Rank is defined as in Table \ref{tab:T5}.}
\label{tab:T7}
\footnotesize
\begin{tabular}{lrrrrrrrr}
\toprule
Model & GSDD & GLID & GLB & SICD & CaFFe & Shelf-Bench & Avg mIoU & Avg Rank \\
\midrule
U-Net & 73.89 & 91.58 & 80.46 & 27.17 & 59.82 & 82.97 & 69.31 & 4.00 \\
GFM-Swin & 73.77 & 83.76 & 75.98 & 30.54 & 57.28 & 86.53 & 67.97 & 6.83 \\
SatlasNet & 76.74 & 85.83 & 77.58 & 31.59 & 52.04 & 83.72 & 67.92 & 4.67 \\
TerraMind & 75.89 & 87.14 & 79.04 & 24.80 & 56.14 & 82.93 & 67.66 & 5.00 \\
DOFA & 77.22 & 84.33 & 79.50 & 13.13 & 56.25 & 84.87 & 65.89 & 5.33 \\
S12-MAE & 75.66 & 85.31 & 77.51 & 28.27 & 43.31 & 82.46 & 65.42 & 7.33 \\
S12-DINO & 76.49 & 85.26 & 77.73 & 21.75 & 43.93 & 82.26 & 64.57 & 7.67 \\
S12-MoCo & 75.51 & 86.08 & 77.92 & 19.03 & 45.11 & 82.30 & 64.33 & 7.33 \\
Prithvi & 75.58 & 84.20 & 77.76 & 25.96 & 39.25 & 80.84 & 63.93 & 9.33 \\
S12-Data2Vec & 76.54 & 86.32 & 77.97 & 12.51 & 43.83 & 84.21 & 63.56 & 6.33 \\
Scale-MAE & 70.38 & 89.47 & 73.16 & 15.18 & 39.38 & 83.96 & 61.92 & 9.83 \\
ViT & 74.41 & 78.14 & 79.81 & 15.58 & 40.25 & 81.86 & 61.67 & 9.67 \\
SpectralGPT & 73.16 & 76.23 & 76.00 & 26.83 & 37.72 & 76.38 & 61.05 & 12.17 \\
CROMA & 75.30 & 72.07 & 77.10 & 16.75 & 39.73 & 82.53 & 60.58 & 10.83 \\
RemoteCLIP & 52.17 & 77.84 & 68.46 & 17.29 & 25.30 & 79.39 & 53.41 & 13.67 \\
\bottomrule
\end{tabular}
\end{table*}

\subsection{Fine-tuning with learning rate optimization} \label{subsec:V-D}
\begin{figure*}[!t]
\centering
\includegraphics[width=0.92\textwidth]{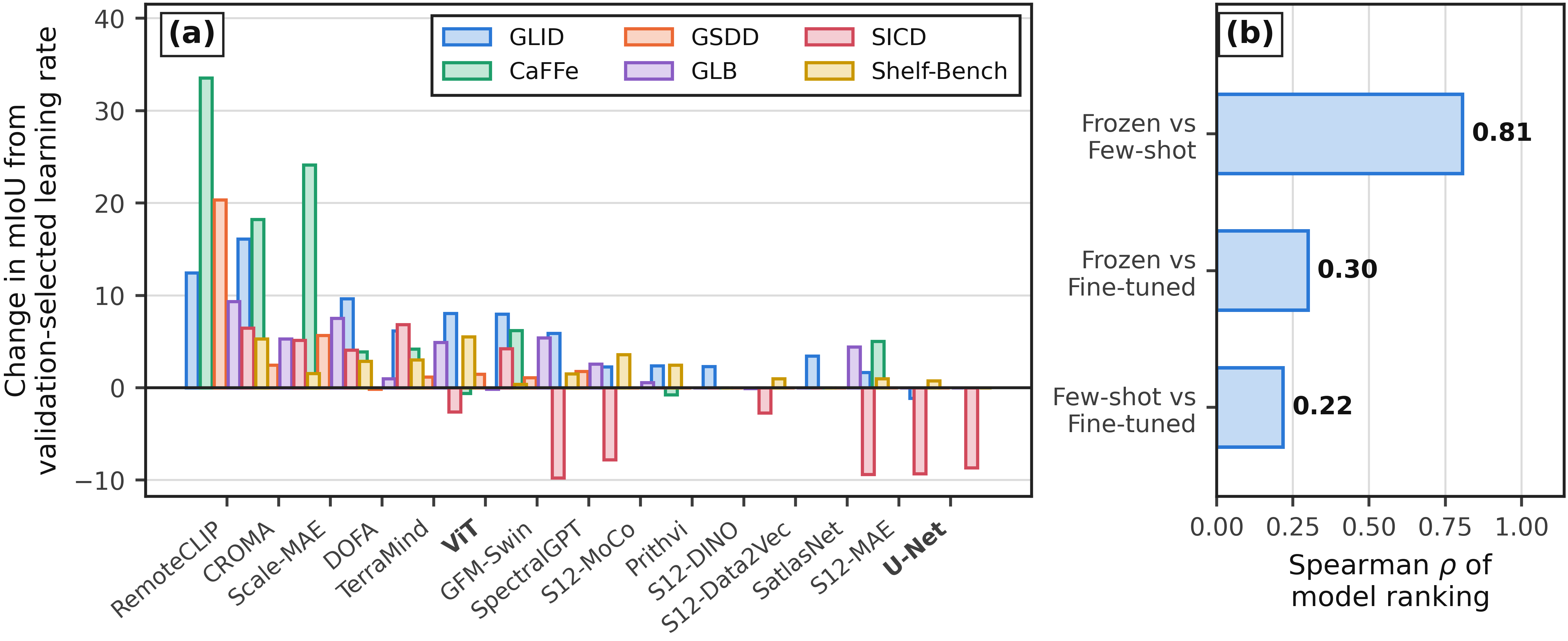}
\caption{Learning rate sensitivity. (a) Change in test mIoU from selecting the learning rate per model, relative to the fixed default; models with no visible bar selected the default. (b) Rank correlation between regimes across the fifteen models.}
\label{fig:F5}
\end{figure*}

The sweep covers all six datasets (Section \ref{sec:4}). Selecting the rate per model on the validation split changes both scores and ordering (Table \ref{tab:T8}, Table \ref{tab:T8b}, Figure \ref{fig:F5}a). On the RGB lake task, DOFA reaches 93.97\% and TerraMind 93.32\%, moving the U-Net to fourth at 90.44\%. On calving fronts, Scale-MAE leads at 63.53\% with GFM-Swin at 63.49\%, moving the U-Net to fifth at 59.82\%. On supraglacial debris, CROMA leads at 77.74\% with DOFA and TerraMind tied at 77.08\%; the U-Net is fourteenth of fifteen at 73.89\%, ahead only of RemoteCLIP.

The size of the gain varies significantly between models, which is precisely why the fixed-rate comparison misleads (Table \ref{tab:T8}). RemoteCLIP gains 33.55 points on calving fronts, 20.35 on supraglacial debris, and Scale-MAE 24.15 on calving fronts, while CROMA gains 16.13 on lakes. Averaged over GFMs, selection is worth 5.42 points on GLID, 7.27 on CaFFe, and 2.49 on GSDD. Thirteen of fifteen models select 10$^{-5}$ on GLID, nine of fifteen on CaFFe, and eight of fifteen on GSDD, so the 10$^{-4}$ default was too large for most encoders on these tasks, and too large by different amounts. The GSDD gain is smaller on average because most models were already close to their best rate at 10$^{-4}$; there, seven of fifteen select 10$^{-4}$ outright, compared with one of fifteen on GLID and six of fifteen on CaFFe but the U-Net still moves: at the fixed rate it beats four GFMs (GFM-Swin, Scale-MAE, SpectralGPT, RemoteCLIP), and under validation-selected rates three of those four overtake it, leaving the U-Net fourteenth of fifteen, ahead only of RemoteCLIP. The same reversal is present on GSDD too, just to a smaller extent.

The second group of datasets mostly continues this pattern. On the multi-source lake task (GLB), TerraMind leads at 83.97\% and CROMA follows at 82.38\%; the U-Net led at the fixed rate (80.46\%) but falls to seventh once the other models are tuned, since its own validation-selected rate is 10$^{-4}$ and its score does not move. RemoteCLIP gains the most of any GFM on GLB, 9.34 points, the same model that shows the largest single gains on calving fronts and supraglacial debris above. On ice-shelf extent, TerraMind again leads, at 88.47\%, with GFM-Swin second at 88.03\%; selection is worth 2.20 points on average across GFMs, and all fifteen models agree between validation- and test-optimal rates, the only dataset in the benchmark where that holds without exception.

Sea ice breaks the pattern. Selection is worth $-$1.46 points on average across GFMs there, the only dataset where validation-based tuning costs performance on average rather than recovering it. The U-Net is the clearest casualty: at the fixed rate it ranks fourth at 27.17\%, but its validation-selected rate of 10$^{-5}$ yields only 18.52\%, an 8.65-point loss that drops it to fourteenth of fifteen, the largest single validation-selection penalty in the benchmark. Only a minority of models gain from selection on sea ice; most are better off, in hindsight, at the fixed rate. Sea ice also has the fewest validation scenes of any dataset in the benchmark (51) and the most severe class imbalance, together with the coarse resolution of input data on which most GFMs have no pretraining; this combination makes it an out-of-distribution problem where most models fail to generalize even with a learning-rate sweep (Table \ref{tab:T8b}). Across all ninety swept model-dataset combinations, validation-based selection matches the test-optimal choice in 73 cases (81\%), reaching 100\% on ice-shelf extent but only 40\% on sea ice, where few validation scenes and severe class imbalance make validation mIoU a noisy selection signal.

\renewcommand{\thetable}{\Roman{table}-A}
\begin{table*}[!t]
\centering
\caption{Fine-tuning with the learning rate selected per model on the validation split; test mIoU is reported for the selected configuration. RGB glacial lake, calving-front and supraglacial debris datasets; continued in Table \ref{tab:T8b} for the multi-source lake, sea-ice and ice-shelf datasets.}
\label{tab:T8}
\scriptsize
\setlength{\tabcolsep}{4pt}
\begin{tabular}{lrrrrrrrrr}
\toprule
& \multicolumn{3}{c}{GLID} & \multicolumn{3}{c}{CaFFe} & \multicolumn{3}{c}{GSDD} \\
\cmidrule(lr){2-4} \cmidrule(lr){5-7} \cmidrule(lr){8-10}
Model & LR & mIoU & vs $10^{-4}$ & LR & mIoU & vs $10^{-4}$ & LR & mIoU & vs $10^{-4}$ \\
\midrule
DOFA & 10$^{-5}$ & 93.97 & +9.63 & 10$^{-5}$ & 60.15 & +3.90 & 10$^{-5}$ & 77.08 & -0.14 \\
TerraMind & 10$^{-5}$ & 93.32 & +6.18 & 10$^{-5}$ & 60.31 & +4.18 & 10$^{-5}$ & 77.08 & +1.19 \\
GFM-Swin & 10$^{-5}$ & 91.75 & +7.98 & 10$^{-5}$ & 63.49 & +6.21 & 10$^{-5}$ & 74.87 & +1.10 \\
U-Net & 10$^{-3}$ & 90.44 & -1.15 & 10$^{-4}$ & 59.82 & +0.00 & 10$^{-4}$ & 73.89 & +0.00 \\
RemoteCLIP & 10$^{-5}$ & 90.28 & +12.44 & 10$^{-5}$ & 58.86 & +33.55 & 10$^{-5}$ & 72.52 & +20.35 \\
Scale-MAE & 10$^{-5}$ & 89.56 & +0.09 & 10$^{-5}$ & 63.53 & +24.15 & 10$^{-5}$ & 76.06 & +5.67 \\
SatlasNet & 10$^{-5}$ & 89.28 & +3.45 & 10$^{-4}$ & 52.04 & +0.00 & 10$^{-4}$ & 76.74 & +0.00 \\
S12-MoCo & 10$^{-5}$ & 88.35 & +2.27 & 10$^{-4}$ & 45.11 & +0.00 & 10$^{-4}$ & 75.51 & +0.00 \\
CROMA & 10$^{-5}$ & 88.20 & +16.13 & 10$^{-5}$ & 57.96 & +18.23 & 10$^{-5}$ & 77.74 & +2.44 \\
S12-DINO & 10$^{-5}$ & 87.58 & +2.32 & 10$^{-4}$ & 43.93 & +0.00 & 10$^{-4}$ & 76.49 & +0.00 \\
S12-MAE & 10$^{-5}$ & 86.95 & +1.65 & 10$^{-5}$ & 48.35 & +5.04 & 10$^{-4}$ & 75.66 & +0.00 \\
Prithvi & 10$^{-5}$ & 86.56 & +2.37 & 10$^{-5}$ & 38.50 & -0.75 & 10$^{-4}$ & 75.58 & +0.00 \\
S12-Data2Vec & 10$^{-4}$ & 86.32 & +0.00 & 10$^{-4}$ & 43.83 & +0.00 & 10$^{-4}$ & 76.54 & +0.00 \\
ViT & 10$^{-5}$ & 86.17 & +8.04 & 10$^{-5}$ & 39.64 & -0.60 & 10$^{-5}$ & 75.90 & +1.49 \\
SpectralGPT & 10$^{-5}$ & 82.13 & +5.90 & 10$^{-4}$ & 37.72 & +0.00 & 10$^{-5}$ & 74.93 & +1.77 \\
\bottomrule
\end{tabular}
\end{table*}

\addtocounter{table}{-1}
\renewcommand{\thetable}{\Roman{table}-B}
\begin{table*}[!t]
\centering
\caption{Fine-tuning with the learning rate selected per model on the validation split; test mIoU is reported for the selected configuration. Multi-source lake, sea-ice and ice-shelf datasets, continuing Table \ref{tab:T8}.}
\label{tab:T8b}
\scriptsize
\setlength{\tabcolsep}{4pt}
\begin{tabular}{lrrrrrrrrr}
\toprule
& \multicolumn{3}{c}{GLB} & \multicolumn{3}{c}{SICD} & \multicolumn{3}{c}{Shelf-Bench} \\
\cmidrule(lr){2-4} \cmidrule(lr){5-7} \cmidrule(lr){8-10}
Model & LR & mIoU & vs $10^{-4}$ & LR & mIoU & vs $10^{-4}$ & LR & mIoU & vs $10^{-4}$ \\
\midrule
TerraMind & 10$^{-5}$ & 83.97 & +4.93 & 10$^{-5}$ & 22.19 & -2.61 & 10$^{-5}$ & 88.47 & +5.54 \\
CROMA & 10$^{-5}$ & 82.38 & +5.28 & 10$^{-5}$ & 21.88 & +5.13 & 10$^{-5}$ & 84.09 & +1.55 \\
SatlasNet & 10$^{-5}$ & 82.01 & +4.42 & 10$^{-5}$ & 22.22 & -9.37 & 10$^{-5}$ & 84.72 & +0.99 \\
GFM-Swin & 10$^{-5}$ & 81.38 & +5.40 & 10$^{-5}$ & 20.79 & -9.75 & 10$^{-5}$ & 88.03 & +1.50 \\
Scale-MAE & 10$^{-5}$ & 80.69 & +7.53 & 10$^{-5}$ & 19.25 & +4.07 & 10$^{-5}$ & 86.85 & +2.89 \\
DOFA & 10$^{-5}$ & 80.47 & +0.97 & 10$^{-5}$ & 20.00 & +6.86 & 10$^{-5}$ & 87.91 & +3.03 \\
U-Net & 10$^{-4}$ & 80.46 & +0.00 & 10$^{-5}$ & 18.52 & -8.65 & 10$^{-4}$ & 82.97 & +0.00 \\
ViT & 10$^{-5}$ & 79.66 & -0.14 & 10$^{-5}$ & 19.83 & +4.25 & 10$^{-5}$ & 82.22 & +0.36 \\
SpectralGPT & 10$^{-5}$ & 78.57 & +2.57 & 10$^{-5}$ & 19.02 & -7.80 & 10$^{-5}$ & 79.99 & +3.60 \\
S12-MoCo & 10$^{-5}$ & 78.49 & +0.56 & 10$^{-4}$ & 19.03 & +0.00 & 10$^{-5}$ & 84.74 & +2.44 \\
S12-Data2Vec & 10$^{-4}$ & 77.97 & +0.00 & 10$^{-4}$ & 12.51 & +0.00 & 10$^{-4}$ & 84.21 & +0.00 \\
RemoteCLIP & 10$^{-5}$ & 77.79 & +9.34 & 10$^{-5}$ & 23.77 & +6.48 & 10$^{-5}$ & 84.68 & +5.29 \\
Prithvi & 10$^{-4}$ & 77.76 & +0.00 & 10$^{-4}$ & 25.96 & +0.00 & 10$^{-4}$ & 80.84 & +0.00 \\
S12-DINO & 10$^{-5}$ & 77.66 & -0.07 & 10$^{-5}$ & 19.00 & -2.74 & 10$^{-5}$ & 83.27 & +1.00 \\
S12-MAE & 10$^{-5}$ & 77.53 & +0.01 & 10$^{-5}$ & 18.98 & -9.29 & 10$^{-5}$ & 83.22 & +0.76 \\
\bottomrule
\end{tabular}
\end{table*}

\addtocounter{table}{-1}
\renewcommand{\thetable}{\Roman{table}-C}
\begin{table*}[!t]
\centering
\caption{Six-dataset average mIoU with the learning rate selected per model on validation (Tables \ref{tab:T8}, \ref{tab:T8b}), compared with the from-scratch U-Net baseline.}
\label{tab:T8c}
\footnotesize
\begin{tabular}{lrr}
\toprule
Model & Avg mIoU & vs U-Net \\
\midrule
TerraMind & 70.89 & +1.58 \\
GFM-Swin & 70.05 & +0.74 \\
DOFA & 69.93 & +0.62 \\
Scale-MAE & 69.32 & +0.01 \\
\textbf{U-Net} & \textbf{69.31} & \textbf{+0.00} \\
CROMA & 68.71 & -0.61 \\
RemoteCLIP & 67.98 & -1.33 \\
SatlasNet & 67.83 & -1.48 \\
S12-MoCo & 65.20 & -4.11 \\
S12-MAE & 65.12 & -4.20 \\
S12-DINO & 64.66 & -4.66 \\
Prithvi & 64.20 & -5.11 \\
ViT & 63.90 & -5.41 \\
S12-Data2Vec & 63.56 & -5.75 \\
SpectralGPT & 62.06 & -7.25 \\
\bottomrule
\end{tabular}
\end{table*}
\renewcommand{\thetable}{\Roman{table}}

Averaged across all six datasets, with each model fine-tuned at its own validation-selected rate, four of the thirteen GFMs exceed the from-scratch U-Net baseline (69.31\%): TerraMind (70.89\%), GFM-Swin (70.05\%), DOFA (69.93\%), and Scale-MAE (69.32\%) (Table \ref{tab:T8c}). This is a second regime, alongside few-shot learning, where pretraining provides a measurable and consistent advantage over training from scratch, once the comparison is not distorted by evaluating every model at a single fixed learning rate.

\subsection{How much does the ranking depend on the regime?} \label{subsec:5.5}
Rankings are stable between frozen and few-shot evaluation ($\rho$ = 0.81, p < 0.001), but weak and not significant between frozen and fine-tuned ($\rho$ = 0.30, p = 0.28) and between few-shot and fine-tuned ($\rho$ = 0.22, p = 0.44) (Figure \ref{fig:F5}b).

Individual movements are large, moving from the frozen ranking to the fixed-rate fine-tuned ranking (again by Avg mIoU), RemoteCLIP falls from third to fifteenth, SatlasNet climbs from tenth to third, and CROMA falls from seventh to fourteenth. Therefore, a frozen-encoder probe poorly predicts fine-tuned performance but is a suitable screen for the few-shot regime. A single-regime leaderboard, the norm in most GFM benchmarks, will mislead anyone whose deployment uses a different regime.

\subsection{Does the pretraining objective matter?}
\label{subsec:5.6}
\begin{figure*}[!t]
\centering
\includegraphics[width=0.92\textwidth]{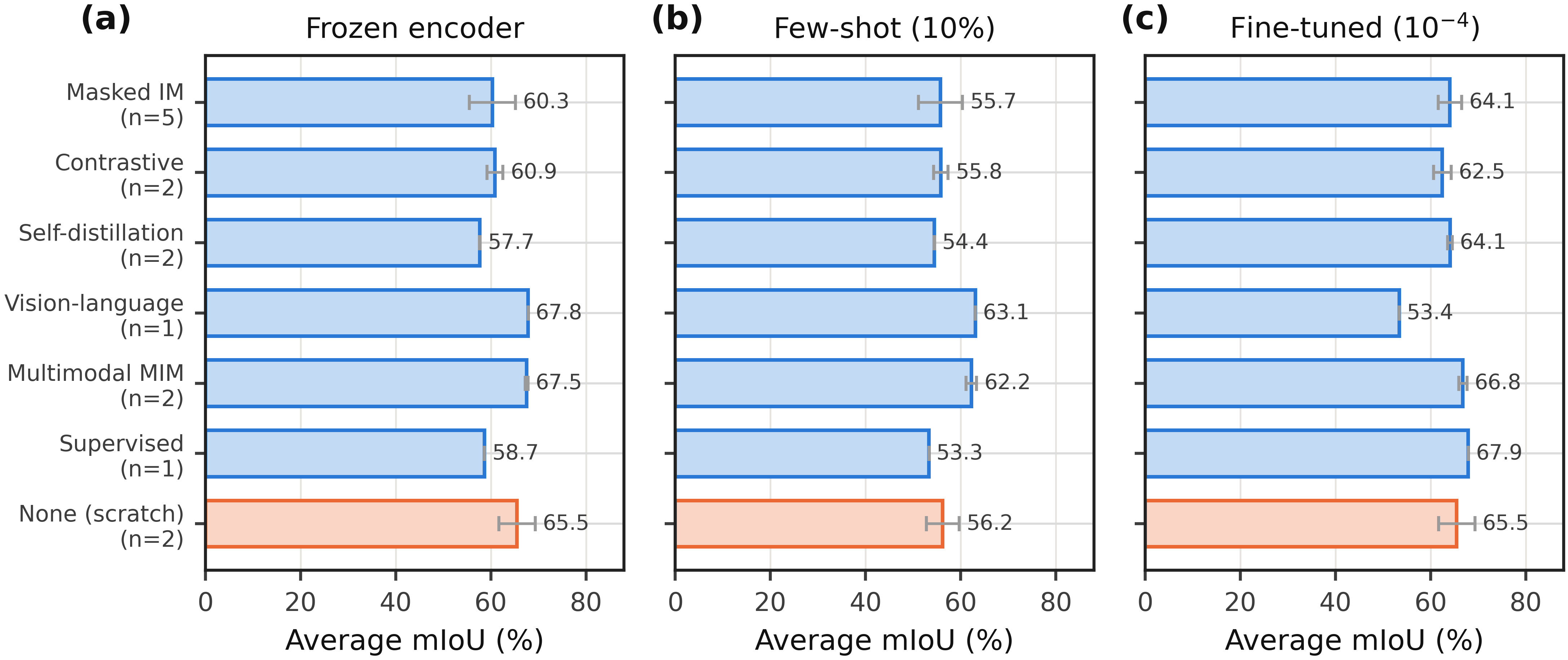}
\caption{Average mIoU by pretraining objective across the three regimes. Error bars are the standard deviation across models within a group; groups with one member have none.}
\label{fig:F6}
\end{figure*}

When encoders are grouped by pretraining objective (Figure \ref{fig:F6}), vision-language pretraining (67.8) and multimodal masked modeling (DOFA, TerraMind: 67.5 ± 0.3) achieve the highest frozen-encoder averages among the pretraining paradigms, though both group means trail the U-Net (69.31). The averages for masked image modeling are 60.3 $\pm$ 4.9, contrastive 60.9 $\pm$ 1.7, self-distillation 57.7 $\pm$ 0.1, and supervised 58.7.

Two observations follow. The masked-modeling group spans a five-point standard deviation despite sharing an objective; DOFA and TerraMind share multimodality but differ in architecture, suggesting that multi-sensor pretraining rather than the masking objective itself is the determining element. Secondly, fine-tuning does not maintain the ordering. Supervised pretraining increases from 58.7 to 67.9, whereas vision-language pretraining decreases from 67.8 frozen to 53.4 fine-tuned, the largest reversal of any group. Objectives that produce strong frozen features are the most sensitive to the fine-tuning learning rate. Since all four members of the SSL4EO-S12 family share the same architecture and pretraining corpus, the goal is neatly isolated. Compared with variation across pretraining corpora, their frozen means are much lower, ranging from 57.6 to 59.2, or 1.6 points, far smaller than the spread across pretraining corpora. What a model was pretrained on is more important on this benchmark than how.

\begin{figure*}[!t]
\centering
\includegraphics[width=0.92\textwidth]{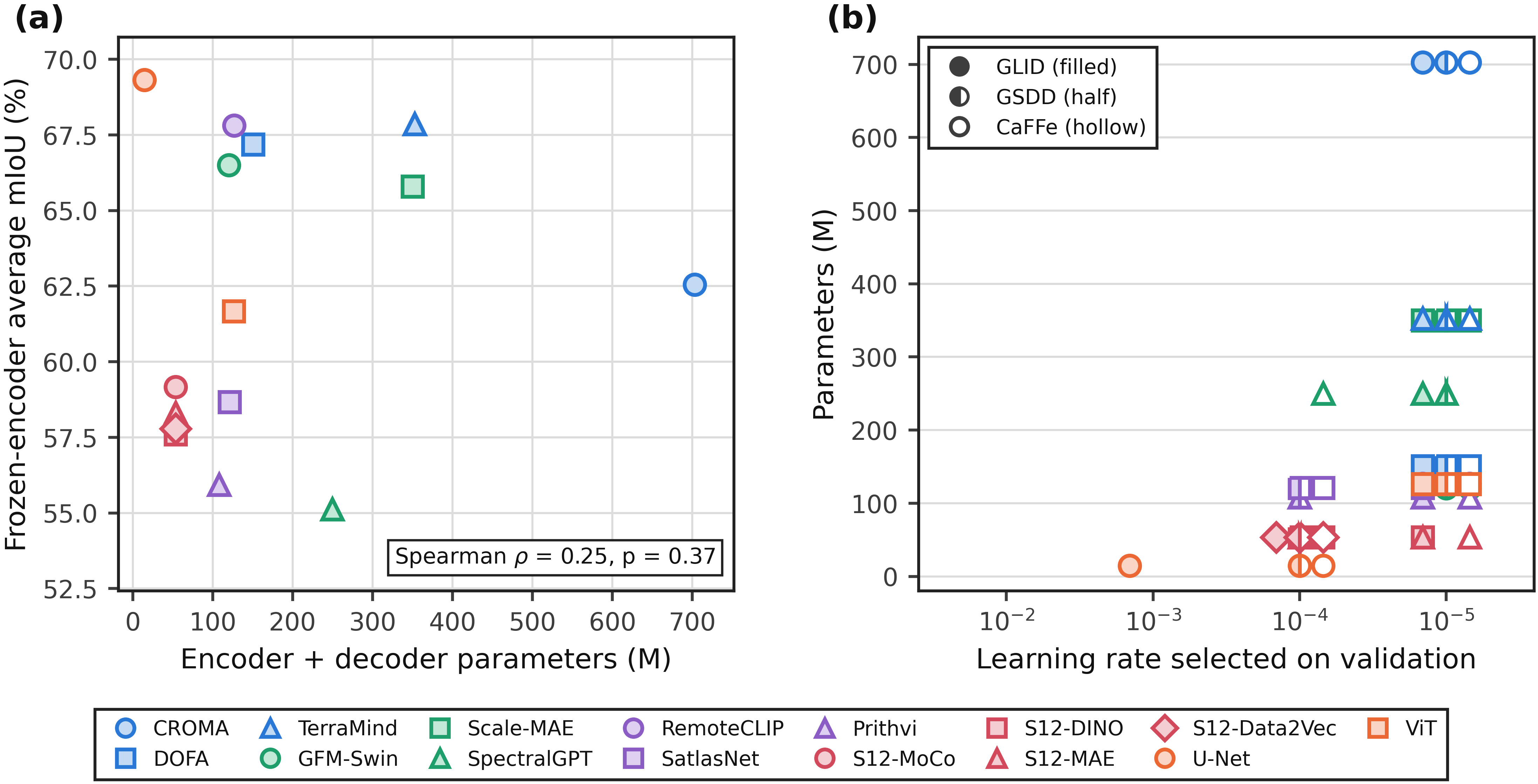}
\caption{Left: parameter count against frozen-encoder accuracy. Right: parameter count against the learning rate selected on validation, for the three swept datasets (GLID filled, GSDD half-filled, CaFFe hollow).}
\label{fig:F7}
\end{figure*}

Though the encoder size contributes little, the correlation between parameter count and frozen accuracy is $\rho$ = +0.25 (p = 0.37) for all the models, and restricting GFMs gives $\rho$ = +0.50 (p = 0.08) (Figure \ref{fig:F7}). With 350 M and 353 M parameters, Scale-MAE and TerraMind rank sixth and second, respectively, while the 14.8 M parameterized U-Net comes in first. Additionally, we find no clear relationship between encoder size and the learning rate selected on validation. These observations do not support the hypothesis that larger models systematically prefer smaller rates: models selecting 10$^{-5}$ span the full size range.

\subsection{Does radar pretraining help on radar tasks?}
\label{subsec:5.7}
\begin{figure}[!t]
\centering
\includegraphics[width=\linewidth]{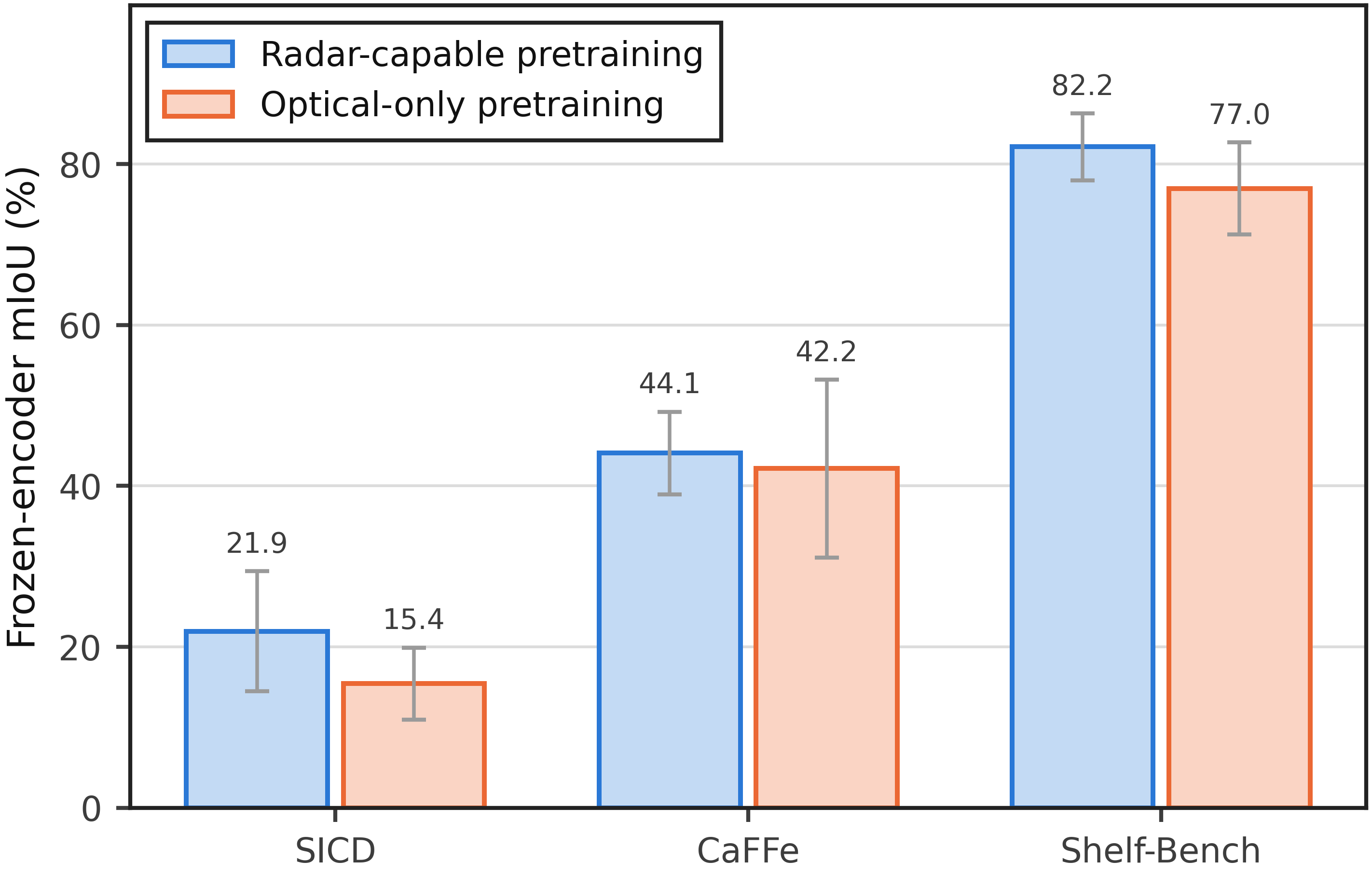}
\caption{Frozen-encoder performance on the three radar tasks, grouped by whether the encoder saw radar during pretraining. Error bars are standard deviations across models.}
\label{fig:F8}
\end{figure}

The benchmark contains three radar-only tasks, providing a direct test of whether radar pretraining transfers better than pretraining without radar (Figure \ref{fig:F8}). We categorize the GFMs into radar-capable (CROMA, DOFA, TerraMind, S12-MAE) and optical-only groups. The radar-capable encoders consistently outperform the optical-only models, achieving a gain of 6.52 points on sea ice (21.94 vs 15.43), 5.19 on ice-shelf extent estimation (82.15 vs 76.96), and 1.94 on calving front detection (44.09 vs 42.15). These results show that including radar data during pretraining offers a modest but quantifiable benefit on downstream radar-based Earth observation tasks. Still, it has less of an effect on the task with the most unique radar signature and is less significant than the performance variability within either model group. The best optical-only encoders either match or outperform radar-capable ones in calving-front mapping. The practical implication is that especially when the target is defined by texture and boundaries rather than backscatter magnitude, an optical-pretrained encoder should not be discounted for a radar task.

\subsection{Per-class analysis} \label{subsec:V-H}
All models perform better on the imbalanced tasks when IoU is averaged across classes. Table \ref{tab:T9} shows that mean foreground IoU is 15.5 points lower than mean mIoU on debris cover, 17.6 points lower on the RGB lake task, and 20.9 points lower on the multi-source lake task. The gap is 1.2 points on the near-balanced ice-shelf task. A single average score is insufficient for cryospheric segmentation because the gap tracks foreground fraction, as expected.
\begin{table*}[!h]
\centering
\caption{Per-class IoU on the sea-ice task with frozen encoders and
full-scene evaluation.}
\label{tab:TS2}
\footnotesize
\setlength{\tabcolsep}{4pt}

\begin{tabular}{lrrrrrrr}
\toprule
Model &
Open water &
New ice &
Young ice &
Thin FY ice &
Thick FY ice &
Old ice &
mIoU \\
\midrule
TerraMind      & 77.82 & 5.04 & 3.13 & 5.39 & 36.35 & 71.87 & 33.27 \\
U-Net          & 68.90 & 0.00 & 0.00 & 0.00 & 29.47 & 64.64 & 27.17 \\
RemoteCLIP     & 64.15 & 0.00 & 0.00 & 0.00 & 34.70 & 46.62 & 24.25 \\
CROMA          & 68.24 & 1.73 & 2.92 & 3.90 & 20.83 & 33.27 & 21.81 \\
DOFA           & 73.44 & 3.92 & 2.99 & 5.44 & 19.28 & 17.36 & 20.41 \\
S12-MoCo       & 61.12 & 0.00 & 0.00 & 0.00 & 19.46 & 38.56 & 19.86 \\
SatlasNet      & 68.88 & 0.00 & 0.00 & 0.00 & 19.72 & 29.12 & 19.62 \\
ViT            & 46.33 & 0.00 & 0.00 & 0.00 & 24.44 & 22.69 & 15.58 \\
SpectralGPT    & 43.24 & 0.00 & 0.00 & 0.00 & 27.68 & 15.43 & 14.39 \\
GFM-Swin       & 44.54 & 0.00 & 0.00 & 0.00 & 22.96 & 17.49 & 14.16 \\
S12-DINO       & 47.74 & 0.00 & 0.06 & 0.00 & 19.82 & 8.82  & 12.74 \\
Prithvi        & 44.98 & 0.00 & 0.00 & 0.00 & 18.27 & 12.24 & 12.58 \\
S12-MAE        & 45.82 & 0.00 & 0.00 & 0.00 & 16.15 & 11.72 & 12.28 \\
S12-Data2Vec   & 44.73 & 0.00 & 0.00 & 0.00 & 16.36 & 7.85  & 11.49 \\
Scale-MAE      & 37.12 & 0.00 & 0.00 & 0.00 & 15.33 & 6.02  & 9.74  \\
\bottomrule
\end{tabular}
\end{table*}

\begin{table}[!h]
\centering
\caption{Frozen-encoder sea-ice results under unweighted and
frequency-weighted aggregation, ordered by weighted IoU.}
\label{tab:TS3}
\footnotesize
\setlength{\tabcolsep}{4pt}

\begin{tabular}{lrrr}
\toprule
Model & mIoU (\%) & Weighted IoU (\%) & Change (pts) \\
\midrule
TerraMind     & 33.27 & 64.64 & +31.38 \\
U-Net         & 27.17 & 57.07 & +29.90 \\
RemoteCLIP    & 24.25 & 48.04 & +23.79 \\
CROMA         & 21.81 & 41.02 & +19.20 \\
S12-MoCo      & 19.86 & 40.80 & +20.95 \\
SatlasNet     & 19.62 & 38.97 & +19.35 \\
DOFA          & 20.41 & 35.10 & +14.70 \\
ViT           & 15.58 & 29.50 & +13.92 \\
GFM-Swin      & 14.16 & 26.26 & +12.10 \\
SpectralGPT   & 14.39 & 25.67 & +11.27 \\
Prithvi       & 12.58 & 23.19 & +10.61 \\
S12-MAE       & 12.28 & 22.87 & +10.59 \\
S12-DINO      & 12.74 & 22.75 & +10.01 \\
S12-Data2Vec  & 11.49 & 20.76 & +9.27  \\
Scale-MAE     & 9.74  & 17.29 & +7.54  \\
\bottomrule
\end{tabular}
\end{table}

\begin{figure}[!t]
\centering
\includegraphics[width=\linewidth]{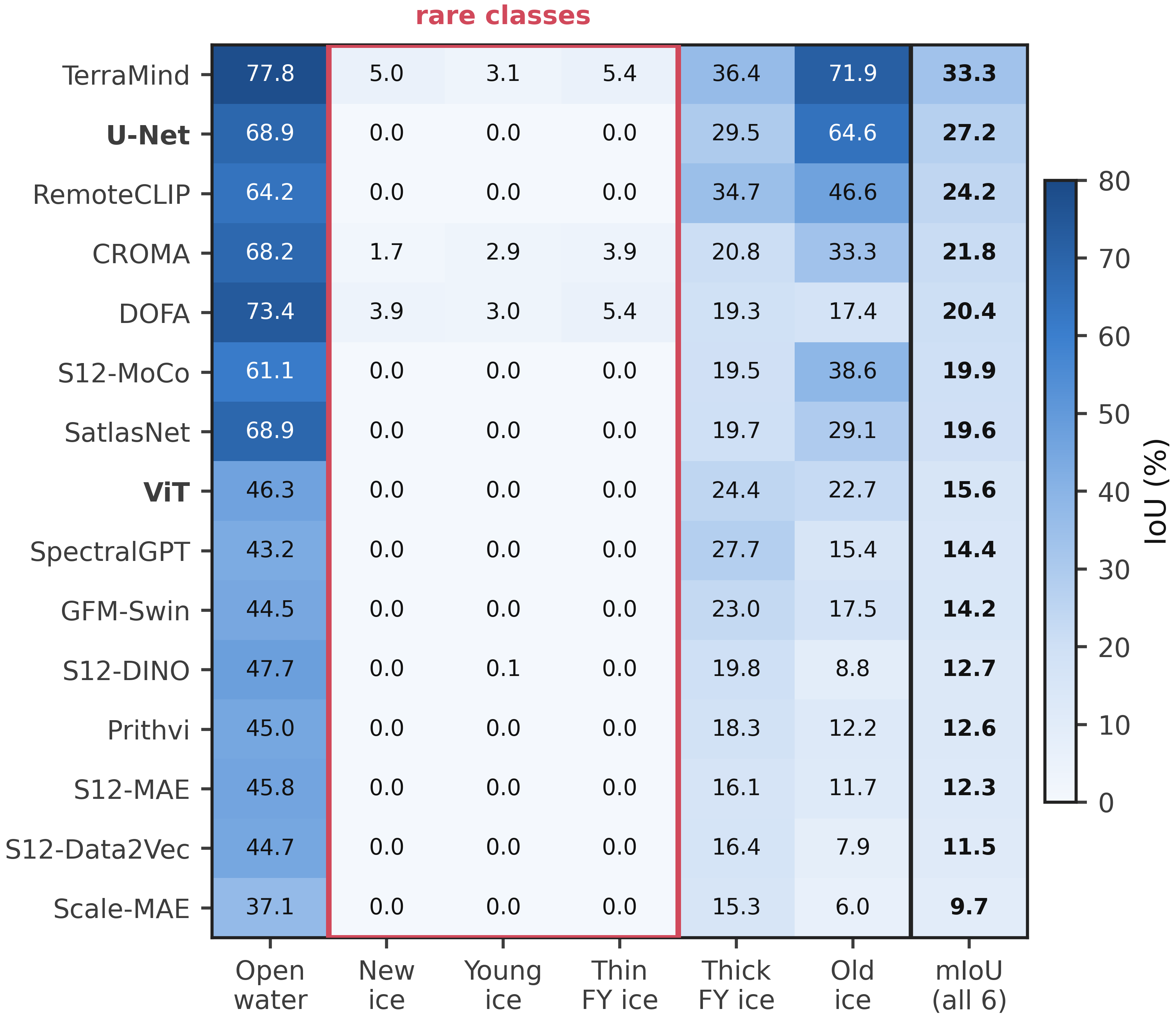}
\caption{Per-class IoU on sea ice, with the reported mIoU, the mean over all six classes, in the final column. The three rare classes are close to zero for every model.}
\label{fig:F9}
\end{figure}

\begin{figure}[!t]
\centering
\includegraphics[width=\linewidth]{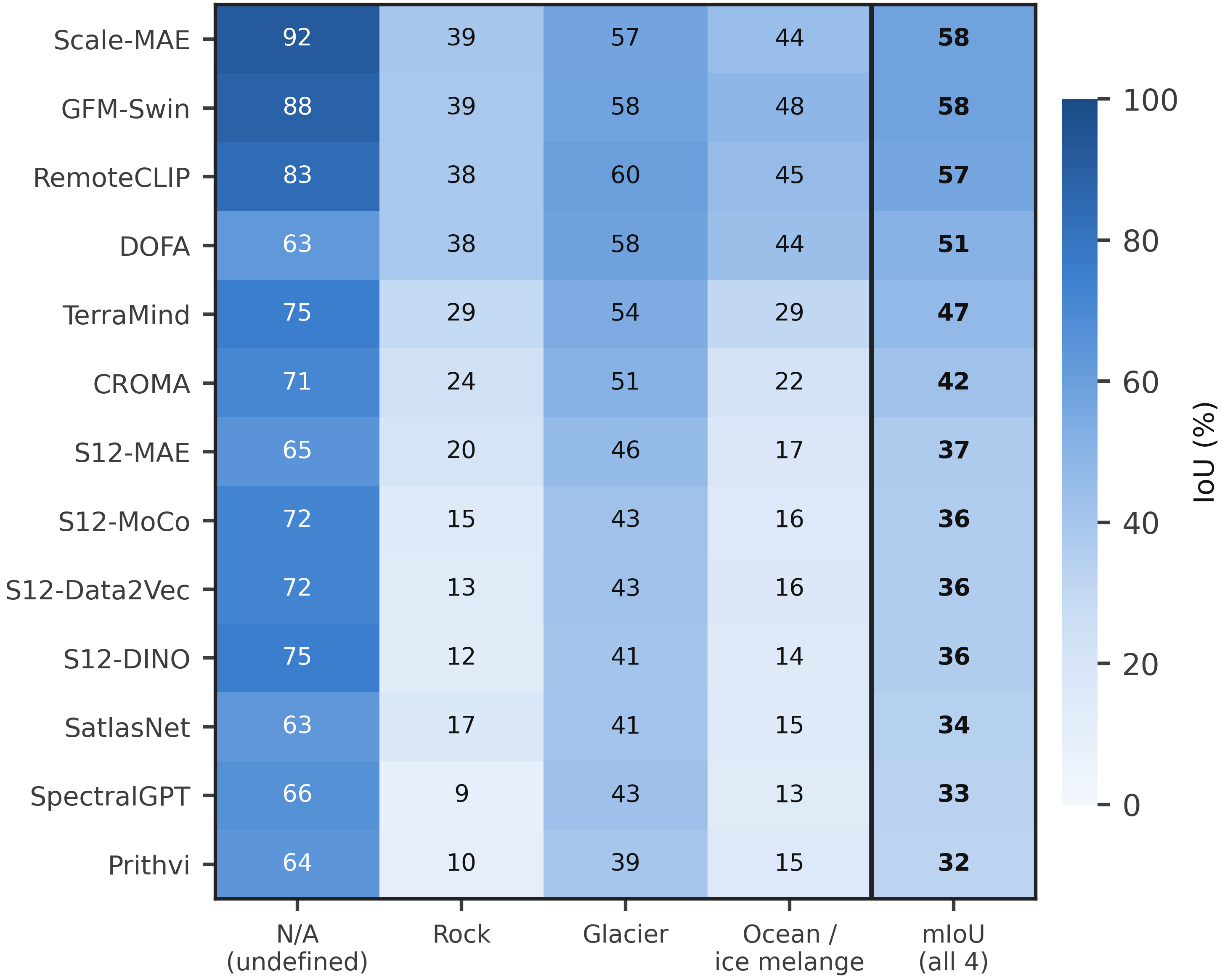}
\caption{Per-class IoU on the four calving-front labels with frozen encoders, with the reported mIoU, the mean over all four, in the final column. The N/A region is not a surface class yet scores highest.}
\label{fig:F10}
\end{figure}

Sea ice shows the same problem in a more serious form (Figure \ref{fig:F9}, Table \ref{tab:TS2}). All models receive reasonable scores on open water, but old-ice scores vary widely, from TerraMind's 71.87\% to Scale-MAE's 6.02\%, and on the three rare classes every model receives almost nothing. Aggregating the six classes significantly affects performance. Following Ice-FMBench \cite{alkaeetaleghan2025}, we provide frequency-weighted averages across the same stage-of-development scheme, weighting each class by its share of ground-truth pixels. Every model is improved by 16.3 points on average and by 31.4 points for TerraMind, whose sea-ice score moves from 33.27\% to 64.64\%, when our frozen findings are re-aggregated in this manner (Table \ref{tab:TS3}). Although the ordering is mostly preserved, the rare classes account for 4.9\% of test pixels, compared with half of the unweighted average. As a result, they contribute less than one percent of the score under the weighted metric, making the failure this section documents invisible: TerraMind would read 64.6\% while segmenting new ice at 5.04\% IoU. For this reason, we report the unweighted mean throughout and provide the weighted values only so results can be compared to benchmarks that aggregate differently. The highest IoU achieved by any model is 5.04\% on new ice, 3.13\% on young ice, and 5.44\% on thin first-year ice, respectively.
In contrast to 5.04\% and 3.13\% on new and young ice, TerraMind's benchmark-leading 33.27\% is based on open water at 77.82\% and old ice at 71.87\%. For the classes that are operationally important for navigation, no model in this study can generate a useful ice chart, and an aggregate of 33\% entirely obscures that. Figure \ref{fig:F10} shows that calving fronts conceal something different. Rock covers 35.7\% but reaches only 22.4\% IoU; ocean with ice melange covers 6.9\% and reaches 24.5\% IoU; and glacier is the biggest and best-segmented of the surface classes at 49.4\% of test pixels, and 48.1\% mean IoU. The highest-scoring label is not a surface class at all: the N/A region, whose boundaries are geometric rather than physical and trivially learned, averages 71.1\% IoU, above every real zone. As the dataset specifies the task and as stated throughout this paper, CaFFe mIoU is the mean across all four labels, meaning that one non-physical label contributes a quarter of the score and raises it by 9.86 points on average relative to the three surface classes alone. This means CaFFe mIoU is not directly comparable to a three-class calving-front score provided elsewhere. Still, it does not affect model comparisons because all models are subject to the same criteria. Per-class results, not the aggregate score, carry the central argument of this section.

\begin{table}[!t]
\centering
\caption{Foreground-class IoU on the four binary tasks with frozen encoders.}
\label{tab:T9}
\footnotesize
\begin{tabular}{lrrrrp{1cm}}
\toprule
Model & GSDD & GLID & GLB & Shelf-Bench & Avg foreground IoU \\
\midrule
DOFA & 57.20 & 85.58 & 60.65 & 86.24 & 72.42 \\
U-Net & 58.77 & 83.56 & 62.83 & 82.05 & 71.80 \\
Scale-MAE & 57.83 & 82.26 & 60.25 & 82.25 & 70.65 \\
TerraMind & 60.11 & 77.11 & 61.61 & 83.31 & 70.53 \\
GFM-Swin & 57.66 & 79.87 & 58.77 & 85.00 & 70.32 \\
RemoteCLIP & 57.95 & 82.22 & 58.12 & 82.59 & 70.22 \\
ViT & 59.58 & 57.40 & 61.43 & 80.75 & 64.79 \\
CROMA & 59.09 & 58.14 & 57.41 & 80.06 & 63.67 \\
S12-MAE & 58.32 & 52.64 & 53.06 & 74.78 & 59.70 \\
SatlasNet & 58.83 & 55.26 & 51.78 & 71.86 & 59.43 \\
S12-MoCo & 57.61 & 52.27 & 53.12 & 73.65 & 59.16 \\
S12-Data2Vec & 58.59 & 51.64 & 51.52 & 74.49 & 59.06 \\
S12-DINO & 54.37 & 52.62 & 52.56 & 74.05 & 58.40 \\
Prithvi & 53.10 & 43.67 & 52.14 & 72.61 & 55.38 \\
SpectralGPT & 57.81 & 43.35 & 48.06 & 64.58 & 53.45 \\
\bottomrule
\end{tabular}
\end{table}

\begin{table}[!h]
\centering
\caption{Frozen-encoder sea-ice mIoU under 8-bit and matched input
standardization, with full-scene evaluation in both cases.}
\label{tab:TS1}
\footnotesize
\setlength{\tabcolsep}{3.5pt}

\begin{tabular}{lrrr}
\toprule
Encoder & 8-bit statistics & Matched statistics & Change (pts) \\
\midrule
S12-MoCo       & 5.76  & 20.66 & +14.90 \\
SatlasNet      & 6.34  & 19.93 & +13.59 \\
U-Net          & 10.07 & 20.61 & +10.54 \\
Prithvi        & 8.07  & 15.37 & +7.30  \\
SpectralGPT    & 5.43  & 12.57 & +7.13  \\
S12-Data2Vec   & 12.74 & 17.25 & +4.50  \\
S12-DINO       & 17.87 & 20.59 & +2.72  \\
GFM-Swin       & 9.27  & 11.55 & +2.28  \\
RemoteCLIP     & 12.55 & 14.51 & +1.97  \\
S12-MAE        & 10.28 & 11.21 & +0.93  \\
ViT            & 13.15 & 12.35 & -0.79  \\
Scale-MAE      & 11.84 & 6.02  & -5.82  \\
\bottomrule
\end{tabular}
\end{table}

\subsection{Sensitivity to input standardization}
\label{subsec:V-I}
Sea-ice inputs encourage a standardization error that is easy to simplify and difficult to detect. When statistics appropriate for 8-bit imaging are used, a mean near 120 is subtracted from data whose values lie between about -4 and +4.5. A direct ablation is obtained by running twelve encoders under both that convention and statistics matched to the data (Table \ref{tab:TS1}); this ablation predates the full-scene 512-pixel retraining used for the other SICD numbers in this paper, so its two columns are internally comparable to each other but not to Table \ref{tab:T5}. Sea-ice mIoU is altered by 4.94 points on average and by as much as 14.90 for S12-MoCo, which goes from 5.76\% to 20.66\%, when the statistics and data are matched. The U-Net gains 10.54 and SatlasNet 13.59. One model moves the other way: under the matched statistics, Scale-MAE loses 5.82.

Two things follow. The impact of input standardization outweighs most model variations reported by this benchmark and is not a detail on tasks whose dynamic range differs from the natural-image conventions many encoders inherit. We also note what we don't do: no encoder uses its own pretraining corpus to get normalization statistics. The magnitude here indicates that per-encoder normalization research is worthwhile, and it maintains a uniform comparison while plausibly disadvantaging frozen backbones.

\subsection{Accuracy against computational cost}
Efficiency comparisons in this setting require careful interpretation. It is inaccurate to attribute the approximately constant GFLOPs of pretrained window-based encoders solely to their fixed patch size, while suggesting that U-Net's computation cost varies primarily with input resolution. Patch size itself is not the determining factor. Instead, windowed encoders incur a relatively fixed computational cost per window, while the cost of segmenting a scene depends on the total number of windows needed to cover the input.

\begin{figure}[!t]
\centering
\includegraphics[width=\linewidth]{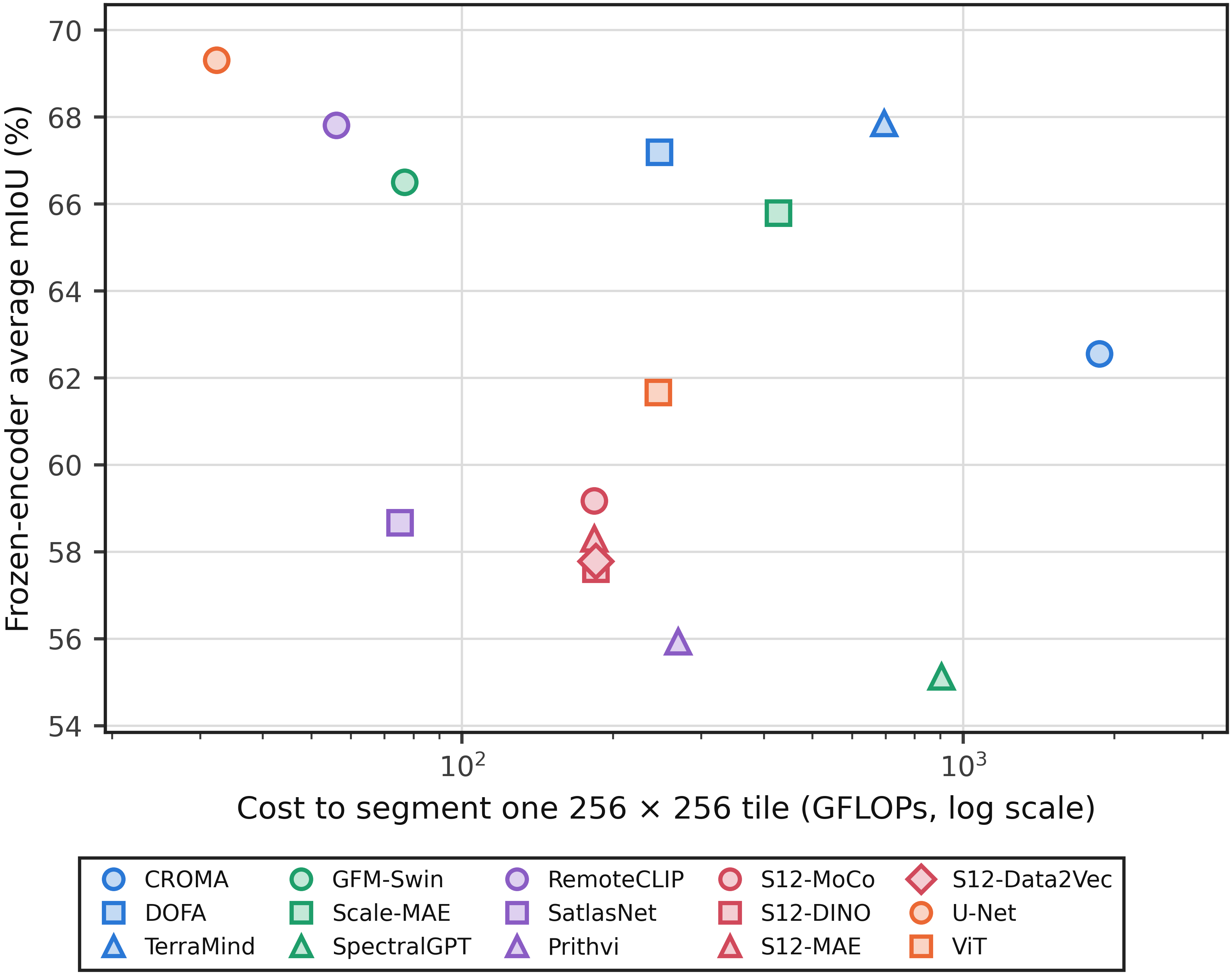}
\caption{Frozen-encoder six-dataset average mIoU versus the computational cost of segmenting one 256 by 256 tile, shown on a logarithmic axis. The frozen regime is used to ensure that the vertical axis is directly comparable with Table \ref{tab:T5} and Figure \ref{fig:F3}.}
\label{fig:F11}
\end{figure}

For a fair comparison, we therefore report computational cost and inference time at a fixed configuration: 256 by 256 pixel tiles, batch size 1, single precision, and the complete encoder and decoder network (Table \ref{tab:T10}, Figure \ref{fig:F11}). The U-Net requires only 32.4 GFLOPs in 1.7 ms and uses 14.8 M parameters to segment a tile, making it the cheapest configuration in the benchmark by a wide margin. CROMA is the most expensive model in the benchmark: its joint optical-SAR configuration carries 702.7 M parameters, and its small 120-pixel window requires nine passes per tile rather than four, so a substantial per-window cost of 207.8 GFLOPs becomes 1870.4 per tile, about 58 $\times$ more than U-Net. SpectralGPT and TerraMind follow at 903.9 and 695.6 GFLOPs per tile, about 28 $\times$ and 21 $\times$ more than U-Net respectively, yet neither improves on it: TerraMind trails by 1.45 points and SpectralGPT by 14.19 points. Thus, per-window cost alone can be misleading; you must also consider the total number of windows required to process a tile.

RemoteCLIP is the least expensive GFM, with 56.1 GFLOPs per tile, requiring less than twice the computation of U-Net. It reaches 67.81\% accuracy when frozen and 90.28\% on the RGB lake task after learning-rate selection. In terms of accuracy per unit of computation, RemoteCLIP ranks second only to the U-Net. For three-band inputs, it achieves competitive accuracy at substantially lower computational cost than much larger encoders.

\begin{figure*}[!t]
\centering
\includegraphics[width=0.92\textwidth]{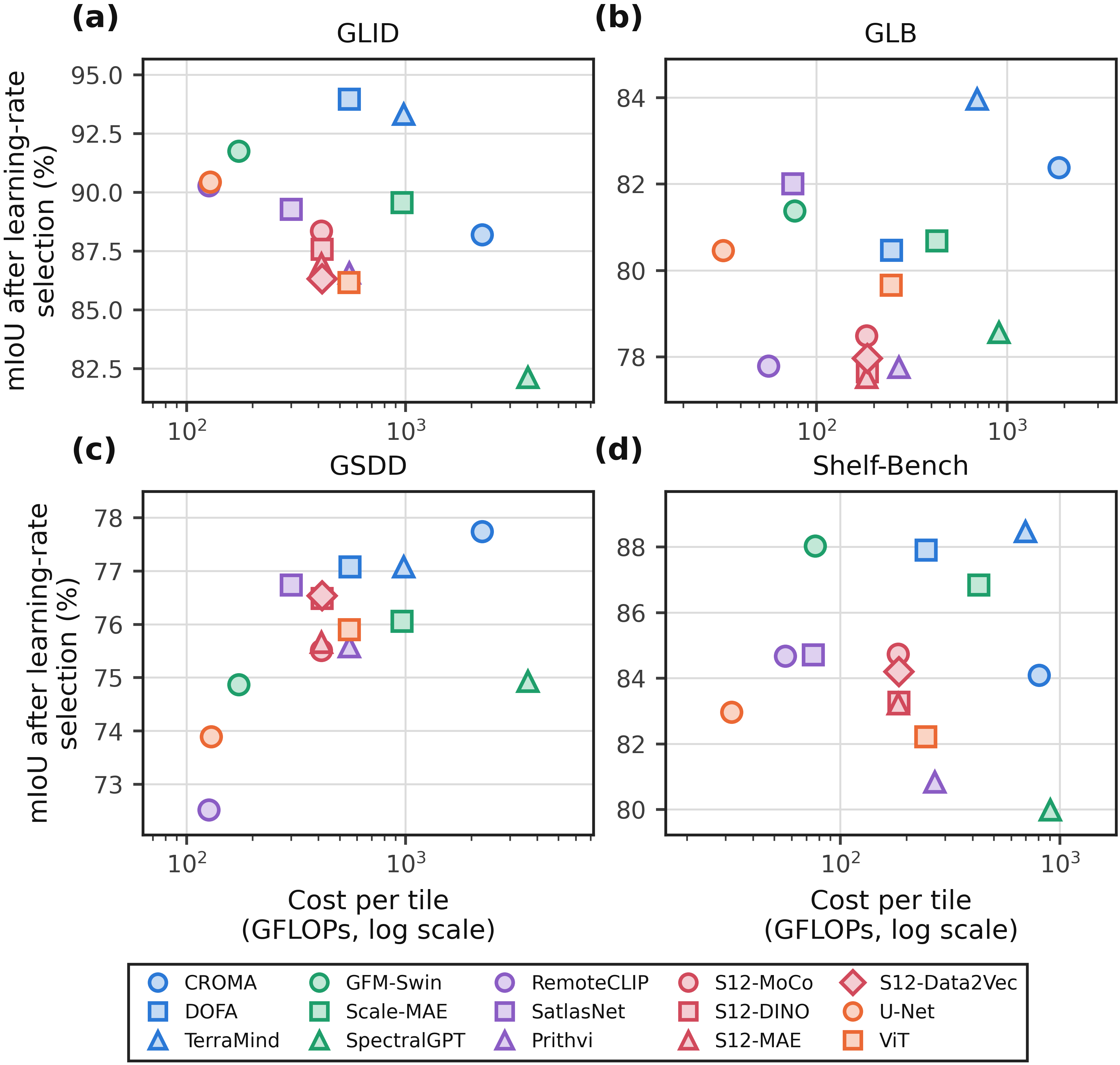}
\caption{Per-dataset accuracy after learning-rate selection plotted against per-tile computational cost, for the four datasets with a fixed native tile size (RGB and multi-source glacial lakes, supraglacial debris, ice-shelf extent). Calving fronts and sea ice are omitted because their tiles are assembled from variable-sized raw scenes, making per-tile cost less directly comparable across models there. Unlike Figure \ref{fig:F11}, the vertical axis represents performance on an individual dataset under tuned fine-tuning, resulting in a different efficiency frontier.}
\label{fig:F12}
\end{figure*}

Figures \ref{fig:F11} and \ref{fig:F12} address complementary questions and should therefore be interpreted together. Figure \ref{fig:F11} compares all models on the frozen six-dataset average, consistent with Table \ref{tab:T5}, and thus reflects overall cost-effectiveness. Figure \ref{fig:F12} instead evaluates a single dataset after per-model learning-rate selection, restricted to the four datasets with a fixed native tile size (the RGB and multi-source lake tasks, supraglacial debris, and ice-shelf extent); calving fronts and sea ice are excluded because their tiles are assembled from variable-sized raw scenes, so a single per-tile cost is not as directly comparable across models on those two as it is on the other four. The resulting efficiency frontier changes accordingly: on the multi-source lake task, GFM-Swin reaches 81.38\% at 76.8 GFLOPs per tile, within 2.6 points of TerraMind's 83.97\% at roughly nine times the cost. This relationship is not evident from the six-dataset average alone.

\begin{table*}[!t]
\centering
\caption{Parameters, per-window and per-tile computational cost, and \ADD{inference} latency, measured on 256 by 256 tiles.}
\label{tab:T10}
\footnotesize
\begin{tabular}{lrrrrrr}
\toprule
Model & Params (M) & Window (px) & GFLOPs/window & Windows/tile & GFLOPs/tile & ms/window \\
\midrule
U-Net & 14.8 & full tile & 32.4 & 1 & 32.4 & 1.7 \\
RemoteCLIP & 126.8 & 224 & 14.0 & 4 & 56.1 & 4.2 \\
SatlasNet & 121.2 & 128 & 18.8 & 4 & 75.1 & 15.3 \\
GFM-Swin & 120.4 & 192 & 19.2 & 4 & 76.8 & 11.6 \\
S12-MAE & 53.5 & 224 & 45.9 & 4 & 183.4 & 4.4 \\
S12-MoCo & 53.5 & 224 & 45.9 & 4 & 183.4 & 4.4 \\
S12-DINO & 53.5 & 224 & 46.2 & 4 & 184.8 & 5.0 \\
S12-Data2Vec & 53.5 & 224 & 46.2 & 4 & 184.8 & 5.9 \\
ViT & 126.3 & 224 & 61.5 & 4 & 246.0 & 4.8 \\
DOFA & 150.7 & 224 & 61.9 & 4 & 247.7 & 6.6 \\
Prithvi & 108.0 & 224 & 67.5 & 4 & 269.8 & 4.2 \\
Scale-MAE & 350.0 & 224 & 106.9 & 4 & 427.4 & 10.6 \\
TerraMind & 352.6 & 224 & 173.9 & 4 & 695.6 & 17.2 \\
SpectralGPT & 249.8 & 128 & 226.0 & 4 & 903.9 & 19.1 \\
CROMA & 702.7 & 120 & 207.8 & 9 & 1870.4 & 23.3 \\
\bottomrule
\end{tabular}
\end{table*}

\subsection{Qualitative behaviour}
\begin{figure*}[!t]
\centering
\includegraphics[width=0.92\textwidth]{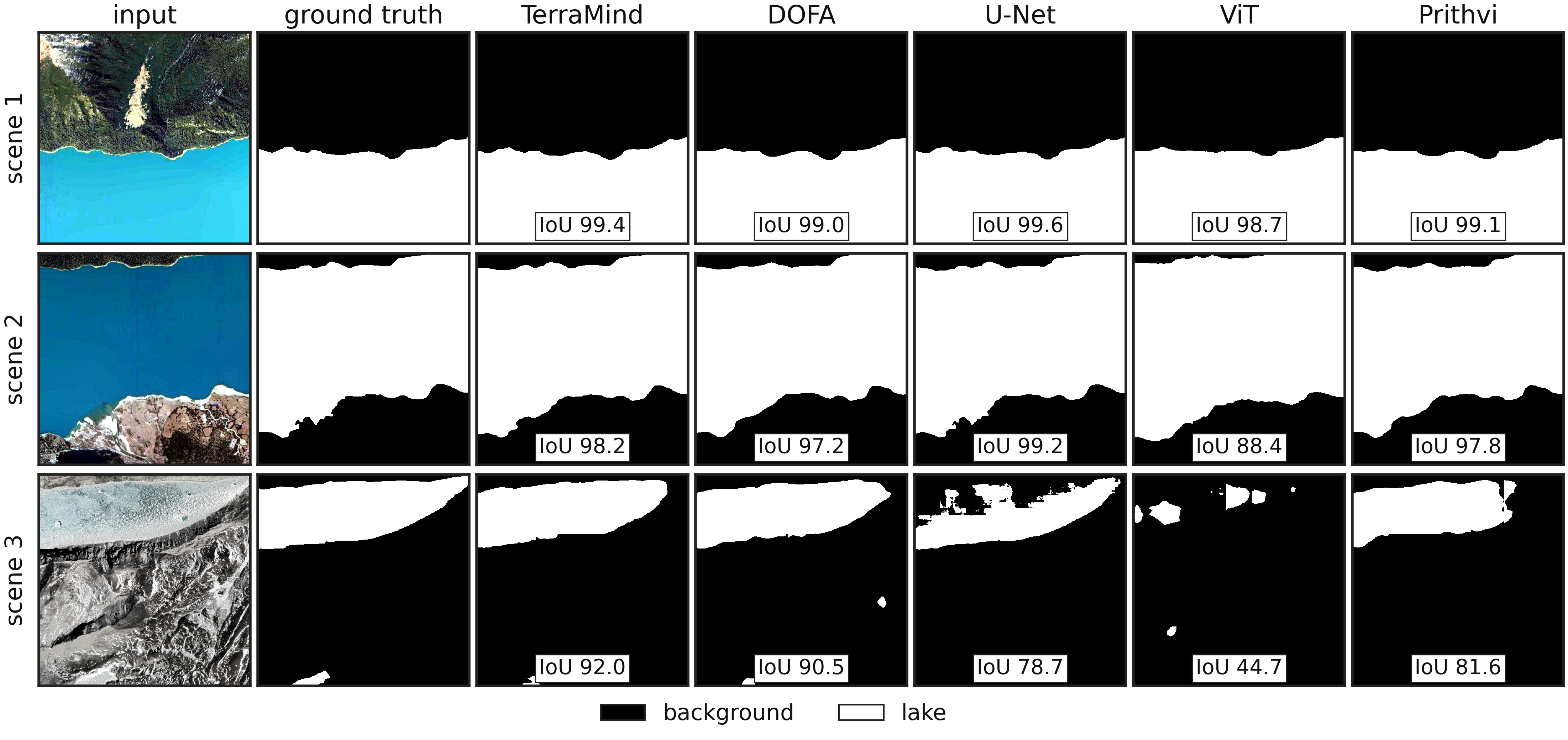}
\caption{Predictions on the RGB glacial lake task (GLID). In the third scene, only TerraMind and DOFA detect the small western lake; U-Net, ViT, and Prithvi all miss it entirely.}
\label{fig:F13}
\end{figure*}

\begin{figure*}[!t]
\centering
\includegraphics[width=0.92\textwidth]{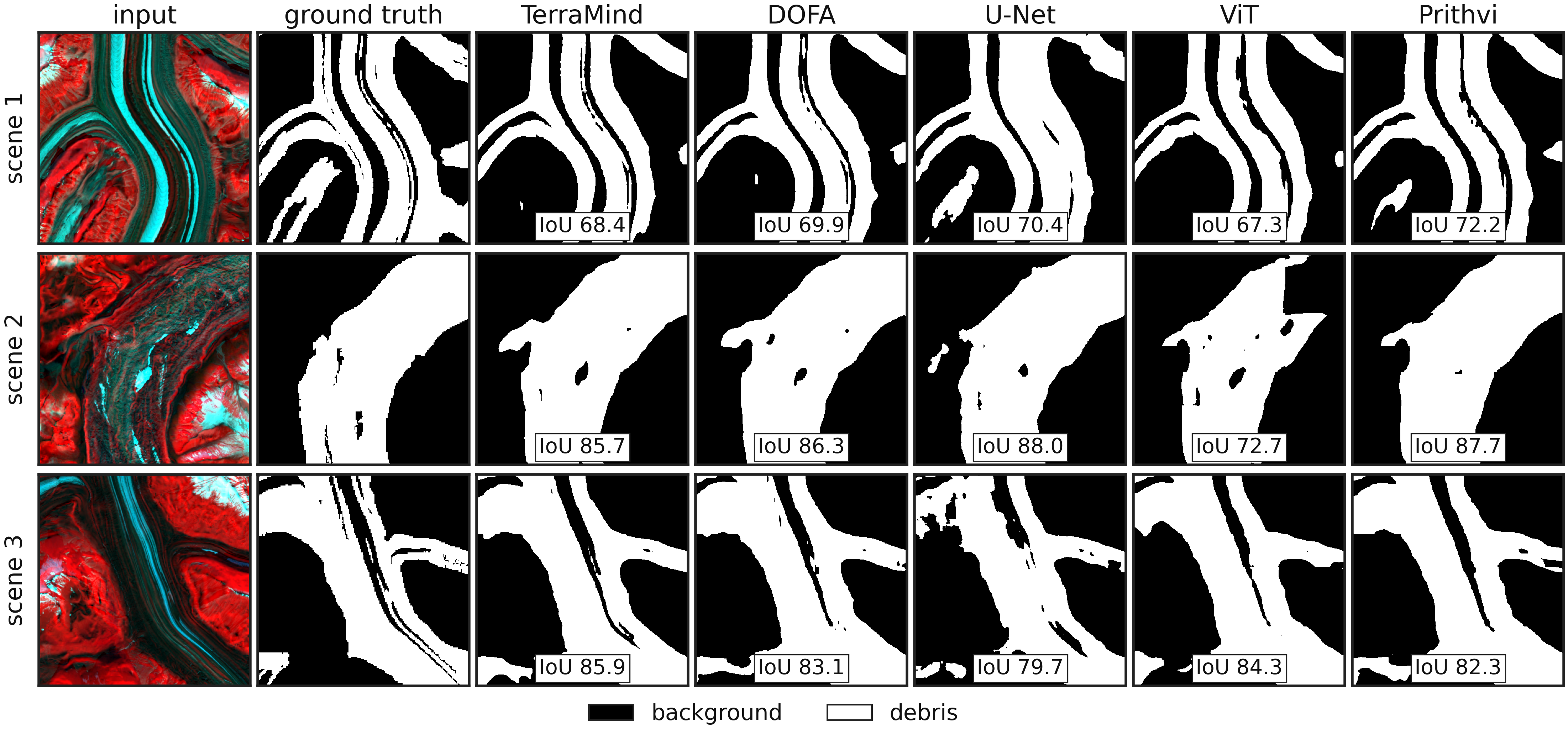}
\caption{Predictions on supraglacial debris (GSDD), the least discriminative task in the benchmark. The input is displayed as a false-color composite (B5, B4, B3), which better distinguishes debris from ice than the natural-color rendering in Figure \ref{fig:F2}.}
\label{fig:F14}
\end{figure*}

\begin{figure*}[!t]
\centering
\includegraphics[width=0.92\textwidth]{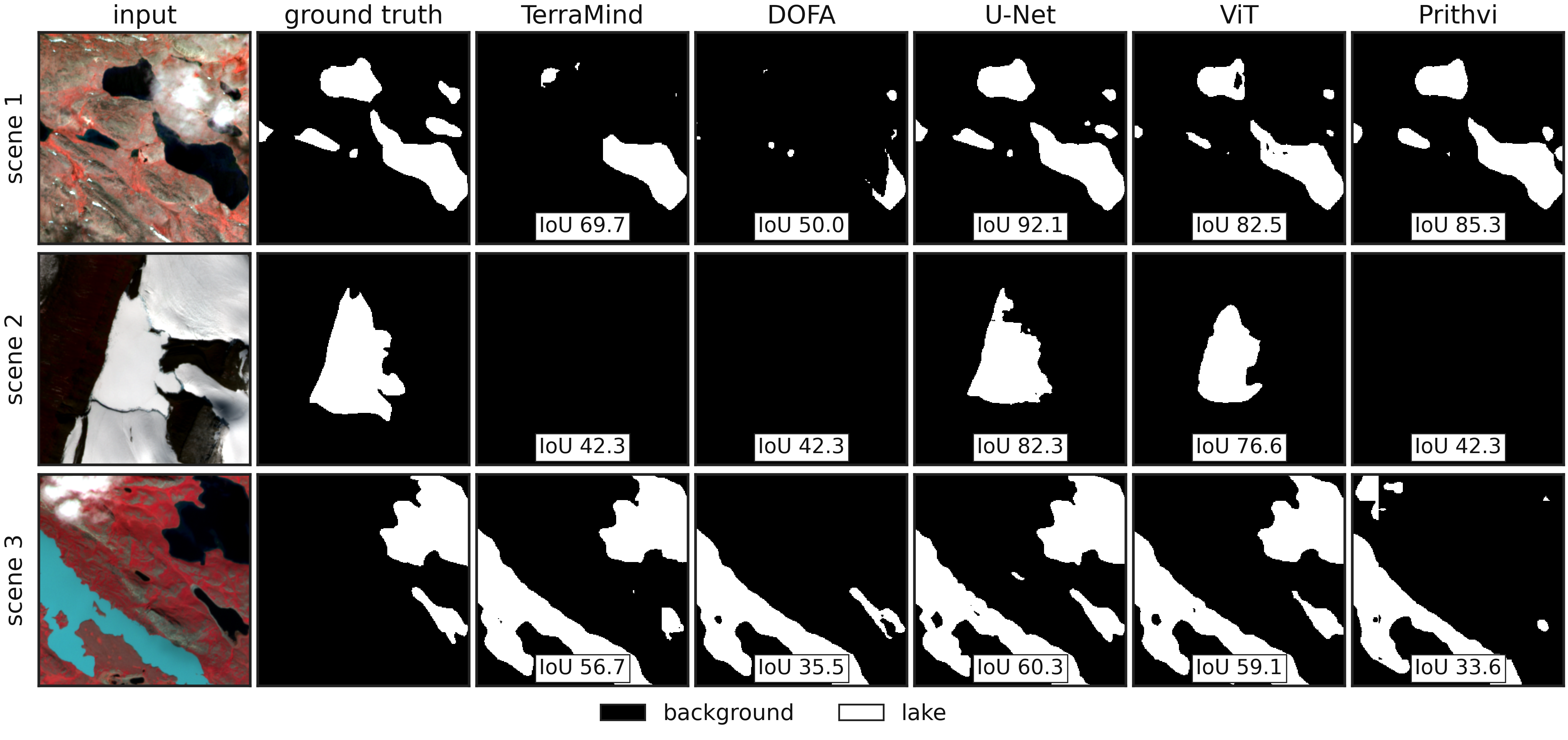}
\caption{Predictions on glacial lakes in the multi-source (GLB) configuration. The input is a false-color composite (B8, B4, B3); since GLB does not include B5, B8 substitutes the standard NIR band, causing water to appear near-black due to strong NIR absorption.}
\label{fig:F15}
\end{figure*}

\begin{figure*}[!t]
\centering
\includegraphics[width=0.92\textwidth]{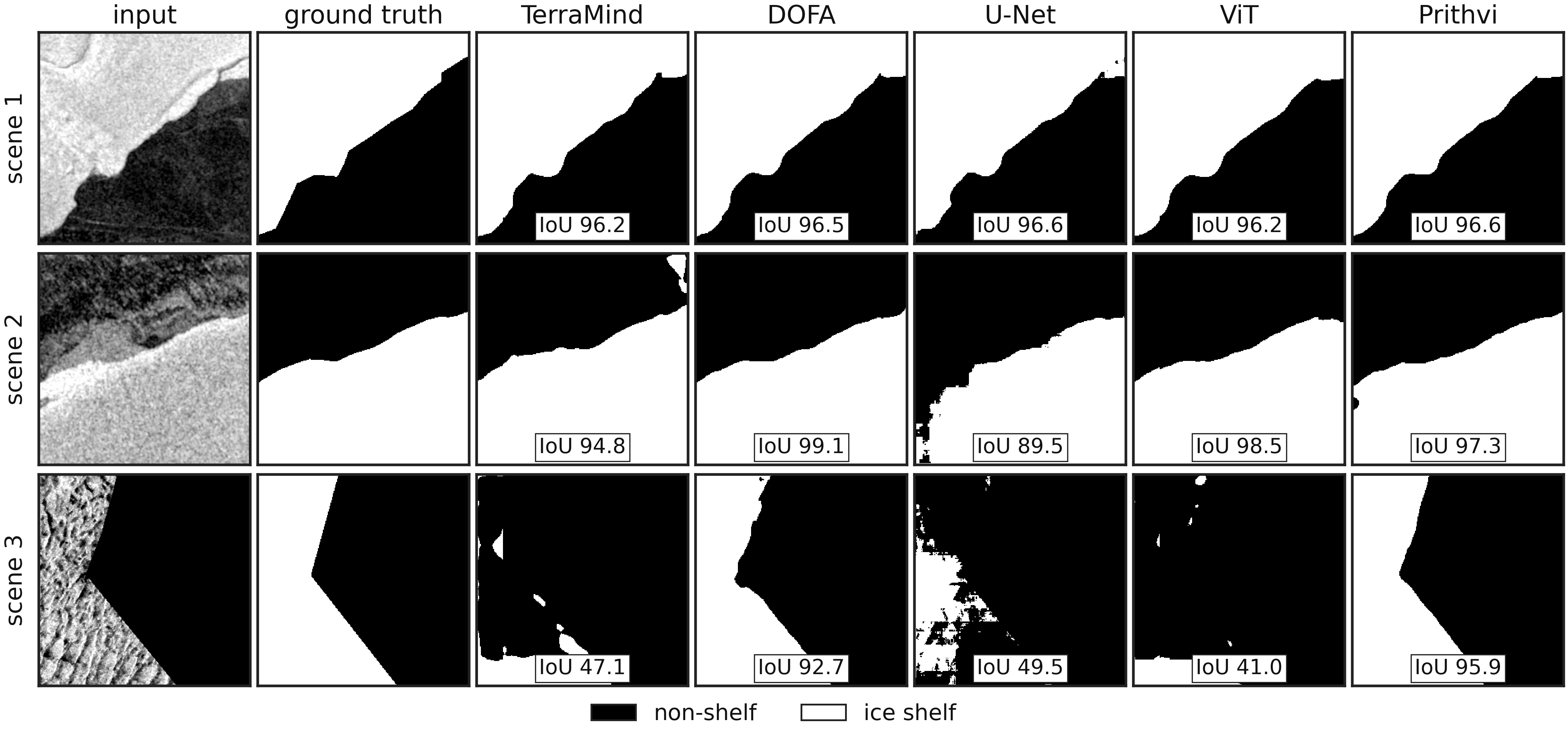}
\caption{Predictions on Antarctic ice-shelf extent (Shelf-Bench).}
\label{fig:F16}
\end{figure*}

\begin{figure*}[!t]
\centering
\includegraphics[width=0.92\textwidth]{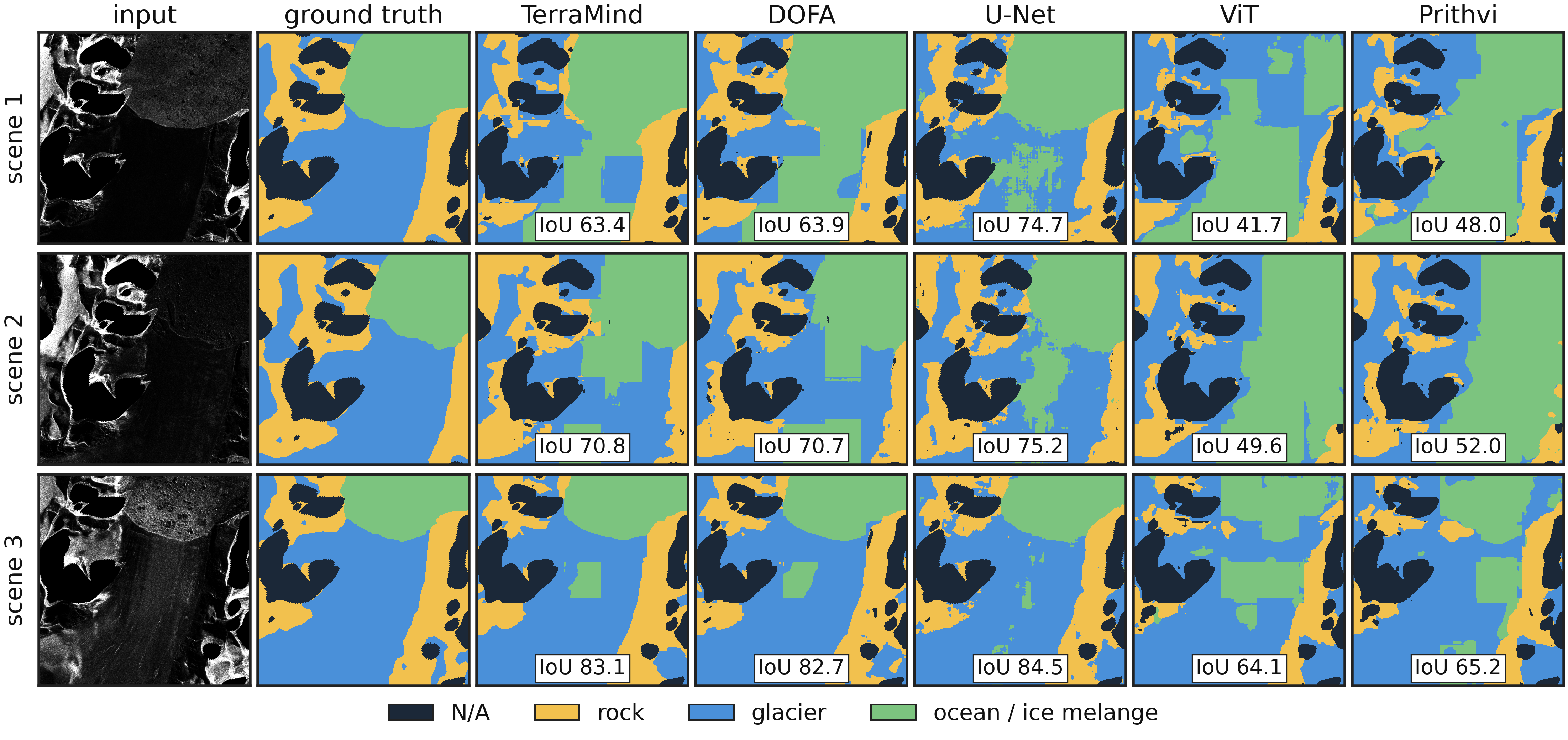}
\caption{Predictions on calving-front zones (CaFFe), coloured by zone. Rectangular discontinuities in the foundation model predictions correspond to sliding-window boundaries.}
\label{fig:F17}
\end{figure*}

\begin{figure*}[!t]
\centering
\includegraphics[width=0.92\textwidth]{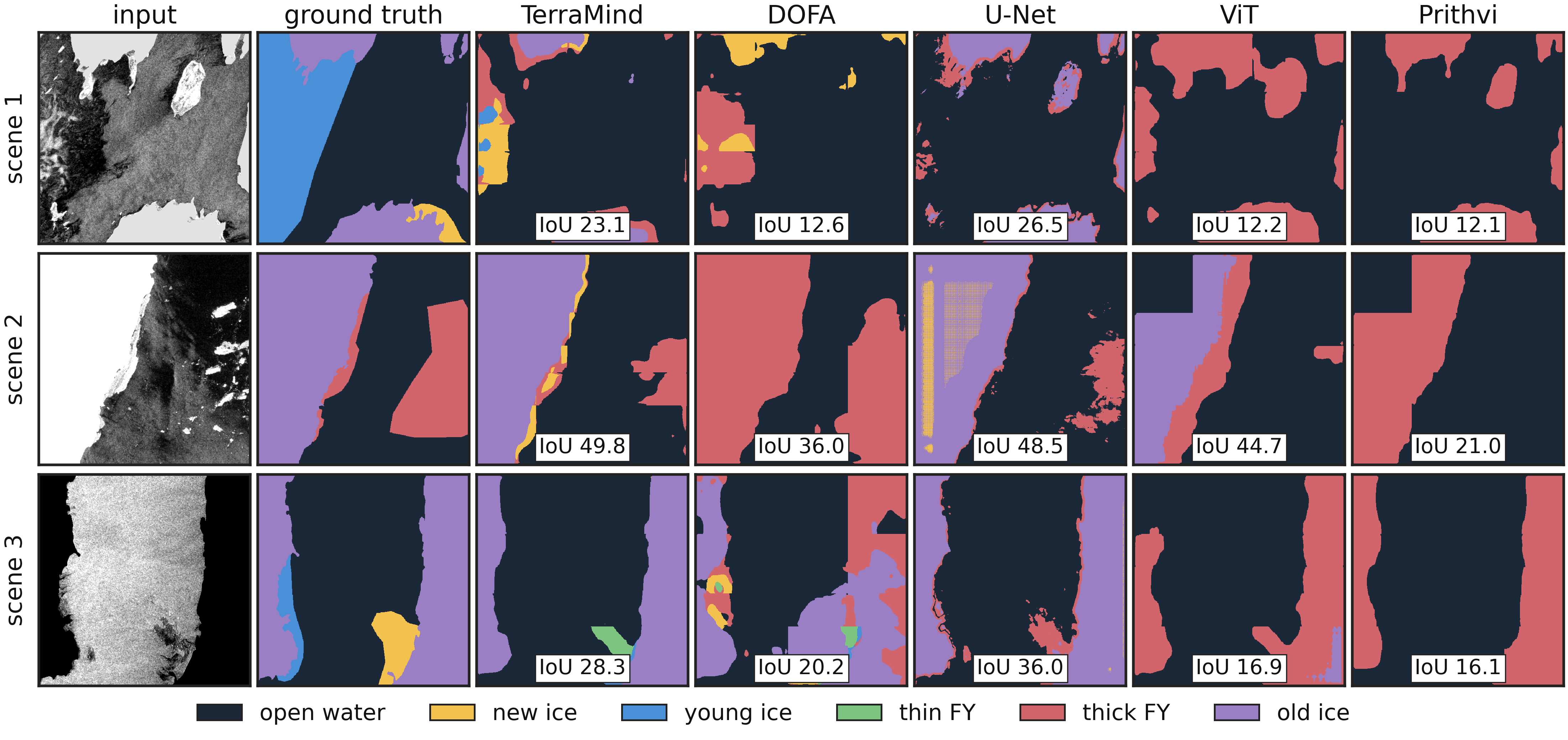}
\caption{Predictions on sea ice (SICD). Prithvi's predictions are confined to open water and thick first-year ice in every scene shown.}
\label{fig:F18}
\end{figure*}

Prediction maps expose failure modes that aggregate metrics may obscure (Figures \ref{fig:F13}-\ref{fig:F18}). Each figure shows the input, the reference labels, and predictions from the same five models, with the per-scene IoU annotated for each prediction. We selected scenes by label diversity rather than foreground fraction, ensuring the visual comparisons meaningfully distinguish model behavior rather than being dominated by largely uniform regions where IoU can become less informative.

Three recurring behaviors are evident. First, across both glacial lake tasks, models agree closely on large water bodies and diverge on small and elongated lakes; in the third GLID scene, TerraMind and DOFA detect a small western lake that the U-Net's full-tile view misses entirely (Figure \ref{fig:F13}). Second, in calving fronts and ice-shelf extent, GFM predictions exhibit faint rectangular discontinuities along sliding-window boundaries, which are absent in the U-Net predicted output (Figure \ref{fig:F17}). This highlights a potential spatial artifact of windowed inference that aggregate metrics may not capture. Third, for sea ice classification, Prithvi's predictions collapse to open water and thick first-year ice only, never predicting any of the other four classes across the three scenes shown (Figure \ref{fig:F18}), consistent with its near-zero per-class IoU on the remaining classes. Although the predictions reasonably distinguish open water from ice, they largely fail to separate individual ice types, explaining the discrepancy between the aggregate score and the per-class results reported in Section \ref{subsec:V-H}.

\section{Discussion}
\subsection{Do foundation models help in the cryosphere?}
The answer depends on the evaluation regime, and the evidence does not support a single, universal ranking. Under frozen encoders with abundant labeled data, we observe no clear advantage for GFMs. A from-scratch U-Net exceeds the best GFM, TerraMind, by 1.45 points, a difference the bootstrap interval confirms is not attributable to test-set composition, while requiring approximately 21$\times$ less inference compute. This suggests that current large-scale pretraining has not yet embedded cryospheric structure in a form that a lightweight decoder can effectively exploit. Given the limited representation of snow and ice in many pretraining corpora, and the lack of explicit distinction between glacial and seasonal ice, this result is not unexpected.

With labeled data being scarce, the performance gap shifts in favor of GFMs. They retain 92.5\% of full-label accuracy, compared with 86.1\% for the U-Net, with five GFMs outright outperforming the baseline. This indicates that pretraining provides a useful prior when supervision is limited, a regime that is particularly relevant to cryospheric applications. With fine-tuning and per-model learning-rate selection, the same reversal is present on five of the six datasets. DOFA and TerraMind outperform U-Net on the RGB lake task, Scale-MAE and GFM-Swin lead on calving fronts, all GFMs except RemoteCLIP outperform U-Net on supraglacial debris, and on the multi-source lake task and ice-shelf extent several GFMs again overtake it (TerraMind, CROMA, SatlasNet, GFM-Swin and Scale-MAE on the former; TerraMind, GFM-Swin, DOFA, Scale-MAE and others on the latter). Thus, the apparent inability of GFMs to surpass a from-scratch baseline is, on these five datasets, partly attributable to evaluating all models at a single learning rate that is not equally suitable for every architecture.

Sea ice is the exception, and for a different reason than the other five. There, validation-based selection actively hurts the U-Net, moving it from fourth at the fixed rate to fourteenth of fifteen, so most models nominally "outperform" it under selection only because its own selected rate performed unusually badly, not because the GFMs themselves improved: sea ice is also where GFM selection is worth $-$1.46 points on average, the only dataset where selection costs performance overall (Section \ref{subsec:V-D}). We do not read the sea-ice ranking under selection as evidence GFMs beat the U-Net there; if anything the fixed-rate comparison is the more trustworthy one for this dataset, given how little validation- and test-optimal rates agree.

\subsection{Factors predicting encoder performance}
Model size does not appear to be a decisive factor. Parameter count shows only a weak and statistically non-significant correlation with frozen accuracy, while the smallest model in the study achieves the strongest average performance. Similarly, the self-supervised learning objective alone does not explain the observed differences: four SSL4EO-S12 encoders with the same architecture and pretraining corpus but different objectives span only 1.6 points. In contrast, pretraining data breadth emerges as the most consistent differentiating factor. Vision-language and multimodal masked-modeling pretraining achieve the two highest frozen-encoder group averages, and radar-capable pretraining provides an advantage across all three radar tasks. Although the effect is modest (2 to 7 points) and smaller than the variations within either group, it represents the clearest structural signal in our analysis. These findings suggest that the pretraining corpus, rather than model size, architecture, or learning objective alone, may offer the greatest opportunity to improve GFM performance in cryospheric applications.

\subsection{Implications of the Benchmark Protocol}
\label{sec:5.10}
Three protocol choices affected performance more than differences across most models, making them particularly important for reproducible evaluation. Learning-rate selection changed GFM performance by an average of 3.2 points across the six datasets, ranging from a 1.46-point loss on sea ice to a 7.27-point gain on calving fronts, with gains of up to 33 points for individual models. Evaluation coverage substantially affected sea-ice results, as evaluating only a center crop covering about 1\% of each scene measures a fundamentally different task than evaluating the full scene. Input standardization also had a pronounced effect, changing sea-ice mIoU by up to 14.9 points. Each of these effects is substantially larger than the 1.45-point difference separating the top two models under frozen-encoder evaluation.

Two further observations reinforce this conclusion. First, model rankings vary across evaluation regimes, indicating that a single leaderboard should not be interpreted as a universal measure of model quality. Second, datasets differ substantially in their ability to distinguish models. For example, the fifteen models span only four percentage points on debris-cover segmentation, meaning that this dataset provides limited discriminatory power while contributing equally to the overall average. These findings suggest that benchmark protocols require the same level of scrutiny as the models being evaluated. A reported ranking can be attributed to encoder quality only when all other aspects of the evaluation are consistently and correctly controlled. The three protocol effects identified above were not apparent from the final results alone; they emerged only when we verified the implementation against the complete evaluation protocol.

\subsection{Guidance for practitioners}
For three-band inputs, including radar used as a proxy, RemoteCLIP offers the best accuracy–efficiency trade-off, ranking second only to the U-Net in accuracy per GFLOP. For multispectral inputs, DOFA delivers the strongest performance at a mid-range computational cost. When computational resources are not a major constraint, or when synthetic modality generation is beneficial, TerraMind is the strongest single encoder and the only model showing a clear advantage on sea ice. When labeled data are abundant, and inference cost is constrained, a from-scratch U-Net remains a strong baseline and should be included in evaluations of new methods. In contrast, when labeled data are scarce, a common scenario in cryospheric mapping, pretrained encoders can justify their additional computational cost, though this benefit is concentrated in a subset of models rather than universal across the roster (Section \ref{subsec:fewshot}). Finally, for rare-class prediction, none of the evaluated models provides consistently adequate performance. Assess these limitations using per-class metrics before deployment, rather than inferring them from aggregate scores alone.

\subsection{Limitations}
We report results using a single random seed. The bootstrap confidence intervals capture sensitivity to test-set composition but do not account for variability from training stochasticity, which may be comparable in magnitude. We therefore consider multi-seed evaluation and corresponding confidence intervals to be the most valuable extension to this benchmark, and we identify them as the priority for future work. When performance differences fall within the reported confidence intervals, we describe the models as statistically indistinguishable rather than assigning a definitive ranking.

The learning-rate sweep covers all six datasets. The fixed-rate choice reduced the GFM's average performance relative to validation-selected rates by 5.42, 7.27 and 2.49 points on the RGB lake task, calving fronts and supraglacial debris, by 3.15 and 2.20 points on the multi-source lake task and ice-shelf extent, and by $-$1.46 points on sea ice, where selection costs rather than recovers performance for the reasons given in Section \ref{subsec:V-D}. We draw fine-tuning conclusions from the validation-selected configuration everywhere except sea ice, where we treat the fixed rate as the more trustworthy comparison given how rarely validation and test selection agree there (6 of 15 models, against at least 80\% on every other dataset). For the frozen and few-shot regimes, we use a single decoder learning rate for all models. Because per-model learning-rate selection has been shown to materially affect fine-tuning performance, it is likely to have a similar impact in these regimes. The corresponding model rankings should therefore be interpreted with this limitation in mind.

No encoder uses its own pretraining normalization, and Section \ref{subsec:V-I} shows normalization choices can materially affect performance. Furthermore, the band and context asymmetries described in Section \ref{subsec:IV-V} favor the baselines on radar and multi-band tasks. Sea ice evaluation deserves separate mention. With only 20 test scenes and three classes occupying approximately one to two percent of pixels, aggregate metrics are subject to substantial variance; consequently, the per-class results provide a more informative assessment of model behavior. In addition, the evaluation uses different tile sizes across encoder groups because of each encoder's native input size, making within-dataset comparisons for sea ice less directly comparable than those for the other datasets.

Finally, the benchmark covers five components but not permafrost, crevasses, or thermokarst, and it is single-date throughout.

\subsection{Future directions}
Our results suggest that cryosphere-specific pretraining may be more promising than relying solely on general Earth-observation models. Cryospheric surfaces exhibit temporal dynamics and physical characteristics that differ substantially from the land-cover classes that dominate existing pretraining datasets. The observed rare-class failures on sea ice further indicate that objectives dominated by frequent classes may be insufficient for generating reliable ice charts. Moreover, the strongest structural signal in our analysis is pretraining-data breadth rather than architecture or model scale. This suggests that future progress is likely to depend primarily on the quality, diversity, and cryosphere-specific coverage of the pretraining data, making data the most important lever for developing more capable foundation models for cryospheric applications.

Three additions could substantially strengthen the benchmark. First, incorporating temporally rich sequences from cold regions would enable encoders to learn seasonal and interannual dynamics rather than relying solely on single-date appearance. Second, adding further cryospheric components, particularly crevasses, permafrost, and thermokarst lakes, would broaden the benchmark’s coverage. Third, incorporating elevation change would provide a more direct measure of cryospheric mass loss, complementing the information captured by surface-area-based assessments.

\section{Conclusions}
Cryo-Bench brings six cryospheric segmentation datasets spanning five components into a unified benchmark, and a fully specified evaluation protocol. We evaluate thirteen geospatial foundation models and compare them with U-Net and ViT baselines under frozen, few-shot, and fine-tuned regimes. 

With frozen encoders, a from-scratch U-Net achieves the highest average (69.31\% mIoU), ahead of TerraMind (67.86\%). The paired difference of 1.45 points has a 95\% confidence interval of [+0.94, +1.98], indicating that the U-Net's lead is statistically resolved. Under a 10\% label regime, foundation models retain 92.5\% of their full-label accuracy, compared with 86.1\% for U-Net, with five models outperforming the baseline, the clearest advantage we measure. With model-specific learning-rate selection, DOFA and TerraMind surpass U-Net on glacial lakes, while Scale-MAE and GFM-Swin outperform it on calving fronts.

Structural analyses show that pretraining-data breadth matters more than the self-supervised objective or model scale. No pretraining paradigm consistently surpasses the from-scratch U-Net baseline under frozen encoders; GFMs gain a clear advantage under limited-label (10\%) supervision, and four of thirteen also surpass U-Net once each model's learning rate is selected on validation and fine-tuned, while four encoders sharing the same architecture and pretraining corpus but differing only in self-supervised objective span only 1.6 points. Moreover, parameter count does not significantly predict accuracy. Per-class evaluation further reveals limitations obscured by aggregate metrics. On imbalanced lake datasets, foreground IoU can be up to 21 points lower than mIoU, while no model achieves more than 5.44\% IoU on any rare sea-ice class, despite a leading aggregate mIoU of 33.27\%. Evaluation protocol choices including learning-rate selection, evaluation coverage, and input standardization also produce larger performance changes than the differences between many models. To support reproducibility and future research, we release Cryo-Bench, the complete results, and the evaluation code for independent verification and extension.

\section*{Data and code availability}
We release the benchmark datasets, per-run logs, the results-collection pipeline, and all evaluation configurations at https://github.com/Sk-2103/Cryo-Bench. For fast and reproducible results, a results dataset and trained model checkpoints are available at https://huggingface.co/datasets/Sk-21/Cryo-Bench.

\section*{Declaration of competing interest}
The authors declare no competing financial interests or personal relationships that could have influenced this work.

\section*{Acknowledgements}

G.C-V acknowledges the support of the European Research Council (ERC) through the USMILE project (grant agreement No.~855187) and the European Commission through the Horizon Europe projects ELIAS (grant agreement No.~101120237) and ELLIOT (grant agreement No.~101214398).

\bibliographystyle{IEEEtran}
\bibliography{refs}



\end{document}